\documentclass[fullname]{clv2-novolnr}

\usepackage{url}
\usepackage[usenames,dvipsnames]{xcolor}
\usepackage{timeline}
\usepackage[normalem]{ulem}
\usepackage{amssymb}
\usepackage{tikz}

\issue{?}{?}{2013}

\newcommand{\todo}[1]{}

\newcommand{\pens}[4]{P\textsuperscript{#1}E\textsuperscript{#2}N\textsuperscript{#3}S\textsuperscript{#4}}
\newcommand{\prop}[1]{\textsc{#1}}

\pgfdeclarelayer{foreground}
\pgfsetlayers{main,foreground}

\newcommand{\pensdiagx}{}
\newcommand{\pensdiagy}{}
\newcommand{\pensdiagn}{}
\newcommand{\pensdiagpa}{}
\newcommand{\pensdiagpb}{}
\newcommand{\pensdiagpc}{}
\newcommand{\pensdiagpd}{}
\newcommand{\pensdiagpe}{}
\newcommand{\pensdiagpf}{}
\newcommand{\pensdiagpg}{}
\newcommand{\pensdiagph}{}
\newcommand{\pensdiagpi}{}
\makeatletter
\define@key{pensdiag}{x}{\renewcommand{\pensdiagx}{#1}}
\define@key{pensdiag}{y}{\renewcommand{\pensdiagy}{#1}}
\define@key{pensdiag}{n}{\renewcommand{\pensdiagn}{#1}}
\define@key{pensdiag}{pa}{\renewcommand{\pensdiagpa}{#1}}
\define@key{pensdiag}{pb}{\renewcommand{\pensdiagpb}{#1}}
\define@key{pensdiag}{pc}{\renewcommand{\pensdiagpc}{#1}}
\define@key{pensdiag}{pd}{\renewcommand{\pensdiagpd}{#1}}
\define@key{pensdiag}{pe}{\renewcommand{\pensdiagpe}{#1}}
\define@key{pensdiag}{pf}{\renewcommand{\pensdiagpf}{#1}}
\define@key{pensdiag}{pg}{\renewcommand{\pensdiagpg}{#1}}
\define@key{pensdiag}{ph}{\renewcommand{\pensdiagph}{#1}}
\define@key{pensdiag}{pi}{\renewcommand{\pensdiagpi}{#1}}
\makeatother

\newenvironment{pensdiagram}[2]{%
\newcommand{\nltoken}[1]{
\node[circle,fill=white,minimum size=3.4mm,draw=black,line width=0.2mm,inner sep=0mm] at (pos) {\scalebox{0.9}{\tiny\color{white}\textsf{\textbf{##1}}}};
}%
\newcommand{\fltoken}[1]{
\node[circle,fill=black,minimum size=3.6mm,draw=white,line width=0.2mm,inner sep=0mm] at (pos) {\scalebox{0.9}{\tiny\color{white}\textsf{\textbf{##1}}}};
}%
\newcommand{\cnltoken}[1]{
\node[circle,fill=blue!70,minimum size=3.6mm,draw=white,line width=0.2mm,inner sep=0mm] at (pos) {\scalebox{0.9}{\tiny\color{white}\textsf{\textbf{##1}}}};
}%
\newcommand{\penstoken}[1]{
\setkeys{pensdiag}{##1}
\ifthenelse{\equal{\pensdiagn}{1}}{
  \node (pos) at (\pensdiagx.5,\pensdiagy.5) {}; \pensdiagpa
}{}
\ifthenelse{\equal{\pensdiagn}{2}}{
  \node (pos) at (\pensdiagx.33,\pensdiagy.67) {}; \pensdiagpa
  \node (pos) at (\pensdiagx.67,\pensdiagy.33) {}; \pensdiagpb
}{}
\ifthenelse{\equal{\pensdiagn}{3}}{
\node (pos) at (\pensdiagx.5,\pensdiagy.7) {}; \pensdiagpa
\node (pos) at (\pensdiagx.28,\pensdiagy.3) {}; \pensdiagpb
\node (pos) at (\pensdiagx.72,\pensdiagy.3) {}; \pensdiagpc
}{}
\ifthenelse{\equal{\pensdiagn}{4}}{
\node (pos) at (\pensdiagx.28,\pensdiagy.72) {}; \pensdiagpa
\node (pos) at (\pensdiagx.72,\pensdiagy.72) {}; \pensdiagpb
\node (pos) at (\pensdiagx.28,\pensdiagy.28) {}; \pensdiagpc
\node (pos) at (\pensdiagx.72,\pensdiagy.28) {}; \pensdiagpd
}{}
\ifthenelse{\equal{\pensdiagn}{5}}{
\node (pos) at (\pensdiagx.23,\pensdiagy.77) {}; \pensdiagpa
\node (pos) at (\pensdiagx.77,\pensdiagy.77) {}; \pensdiagpb
\node (pos) at (\pensdiagx.5,\pensdiagy.5) {}; \pensdiagpc
\node (pos) at (\pensdiagx.23,\pensdiagy.23) {}; \pensdiagpd
\node (pos) at (\pensdiagx.77,\pensdiagy.23) {}; \pensdiagpe
}{}
\ifthenelse{\equal{\pensdiagn}{6}}{
\node (pos) at (\pensdiagx.32,\pensdiagy.79) {}; \pensdiagpa
\node (pos) at (\pensdiagx.68,\pensdiagy.79) {}; \pensdiagpb
\node (pos) at (\pensdiagx.2,\pensdiagy.5) {}; \pensdiagpc
\node (pos) at (\pensdiagx.8,\pensdiagy.5) {}; \pensdiagpd
\node (pos) at (\pensdiagx.32,\pensdiagy.21) {}; \pensdiagpe
\node (pos) at (\pensdiagx.68,\pensdiagy.21) {}; \pensdiagpf
}{}
\ifthenelse{\equal{\pensdiagn}{7}}{
\node (pos) at (\pensdiagx.31,\pensdiagy.79) {}; \pensdiagpa
\node (pos) at (\pensdiagx.69,\pensdiagy.79) {}; \pensdiagpb
\node (pos) at (\pensdiagx.2,\pensdiagy.5) {}; \pensdiagpc
\node (pos) at (\pensdiagx.5,\pensdiagy.5) {}; \pensdiagpd
\node (pos) at (\pensdiagx.8,\pensdiagy.5) {}; \pensdiagpe
\node (pos) at (\pensdiagx.31,\pensdiagy.21) {}; \pensdiagpf
\node (pos) at (\pensdiagx.69,\pensdiagy.21) {}; \pensdiagpg
}{}
\ifthenelse{\equal{\pensdiagn}{8}}{
\node (pos) at (\pensdiagx.2,\pensdiagy.79) {}; \pensdiagpa
\node (pos) at (\pensdiagx.8,\pensdiagy.79) {}; \pensdiagpb
\node (pos) at (\pensdiagx.5,\pensdiagy.65) {}; \pensdiagpc
\node (pos) at (\pensdiagx.2,\pensdiagy.5) {}; \pensdiagpd
\node (pos) at (\pensdiagx.8,\pensdiagy.5) {}; \pensdiagpe
\node (pos) at (\pensdiagx.5,\pensdiagy.35) {}; \pensdiagpf
\node (pos) at (\pensdiagx.2,\pensdiagy.21) {}; \pensdiagpg
\node (pos) at (\pensdiagx.8,\pensdiagy.21) {}; \pensdiagph
}{}
\ifthenelse{\equal{\pensdiagn}{9}}{
\node (pos) at (\pensdiagx.2,\pensdiagy.79) {}; \pensdiagpa
\node (pos) at (\pensdiagx.5,\pensdiagy.79) {}; \pensdiagpb
\node (pos) at (\pensdiagx.8,\pensdiagy.79) {}; \pensdiagpc
\node (pos) at (\pensdiagx.2,\pensdiagy.5) {}; \pensdiagpd
\node (pos) at (\pensdiagx.5,\pensdiagy.5) {}; \pensdiagpe
\node (pos) at (\pensdiagx.8,\pensdiagy.5) {}; \pensdiagpf
\node (pos) at (\pensdiagx.2,\pensdiagy.21) {}; \pensdiagpg
\node (pos) at (\pensdiagx.5,\pensdiagy.21) {}; \pensdiagph
\node (pos) at (\pensdiagx.8,\pensdiagy.21) {}; \pensdiagpi
}{}
}%
\begin{tikzpicture}
[every token/.style={minimum size=3.8mm,draw=white,line width=0.2mm,inner sep=0mm}]
\begin{pgfonlayer}{foreground}
\node[anchor=north] at (3.5,0.7) {\sf #1};
\draw[->,thick] (1,1) -- (6.3,1);
\node[anchor=north] at (1.5,1) {\small\sf 1};
\draw (2,1) -- (2,6);
\node[anchor=north] at (2.5,1) {\small\sf 2};
\draw (3,1) -- (3,6);
\node[anchor=north] at (3.5,1) {\small\sf 3};
\draw (4,1) -- (4,6);
\node[anchor=north] at (4.5,1) {\small\sf 4};
\draw (5,1) -- (5,6);
\node[anchor=north] at (5.5,1) {\small\sf 5};
\draw (6,1) -- (6,6);
\node[anchor=south,rotate=90] at (0.7,3.5) {\sf #2};
\draw[->,thick] (1,1) -- (1,6.3);
\node[anchor=east] at (1,1.5) {\small\sf 1};
\draw (1,2) -- (6,2);
\node[anchor=east] at (1,2.5) {\small\sf 2};
\draw (1,3) -- (6,3);
\node[anchor=east] at (1,3.5) {\small\sf 3};
\draw (1,4) -- (6,4);
\node[anchor=east] at (1,4.5) {\small\sf 4};
\draw (1,5) -- (6,5);
\node[anchor=east] at (1,5.5) {\small\sf 5};
\draw (1,6) -- (6,6);
\end{pgfonlayer}
}{
\end{tikzpicture}%
}

\newcommand{\propct}{0}
\newcommand{\propcf}{0}
\newcommand{\propcw}{0}
\newcommand{\propcs}{0}
\newcommand{\propcd}{0}
\newcommand{\propca}{0}
\newcommand{\propci}{0}
\newcommand{\propcg}{0}
\newcommand{\proptf}{0}
\newcommand{\proptw}{0}
\newcommand{\propts}{0}
\newcommand{\proptd}{0}
\newcommand{\propta}{0}
\newcommand{\propti}{0}
\newcommand{\proptg}{0}
\newcommand{\propfw}{0}
\newcommand{\propfs}{0}
\newcommand{\propfd}{0}
\newcommand{\propfa}{0}
\newcommand{\propfi}{0}
\newcommand{\propfg}{0}
\newcommand{\propws}{0}
\newcommand{\propwd}{0}
\newcommand{\propwa}{0}
\newcommand{\propwi}{0}
\newcommand{\propwg}{0}
\newcommand{\propsd}{0}
\newcommand{\propsa}{0}
\newcommand{\propsi}{0}
\newcommand{\propsg}{0}
\newcommand{\propda}{0}
\newcommand{\propdi}{0}
\newcommand{\propdg}{0}
\newcommand{\propai}{0}
\newcommand{\propag}{0}
\newcommand{\propig}{0}

\newcommand{\numberofcnls}{100}
\newcommand{\propaabs}{50}

\newcommand{\propcabs}{45}

\newcommand{\propdabs}{53}

\newcommand{\propfabs}{54}

\newcommand{\propgabs}{10}

\newcommand{\propiabs}{43}

\newcommand{\propsabs}{7}

\newcommand{\proptabs}{22}

\newcommand{\propwabs}{93}

\renewcommand{\propag}{1}
\renewcommand{\propai}{4}
\renewcommand{\propca}{4}
\renewcommand{\propcd}{33}
\renewcommand{\propcf}{3}
\renewcommand{\propcg}{8}
\renewcommand{\propci}{33}
\renewcommand{\propcs}{6}
\renewcommand{\propct}{17}
\renewcommand{\propcw}{40}
\renewcommand{\propda}{20}
\renewcommand{\propdg}{6}
\renewcommand{\propdi}{29}
\renewcommand{\propfa}{45}
\renewcommand{\propfd}{19}
\renewcommand{\propfg}{2}
\renewcommand{\propfi}{10}
\renewcommand{\propfs}{1}
\renewcommand{\propfw}{52}
\renewcommand{\propsd}{6}
\renewcommand{\propsg}{6}
\renewcommand{\propsi}{1}
\renewcommand{\propta}{5}
\renewcommand{\proptd}{17}
\renewcommand{\proptf}{1}
\renewcommand{\propti}{18}
\renewcommand{\proptw}{21}
\renewcommand{\propwa}{49}
\renewcommand{\propwd}{46}
\renewcommand{\propwg}{5}
\renewcommand{\propwi}{42}
\renewcommand{\propws}{1}
\newcommand{\propaavge}{2.5}
\newcommand{\propcavge}{4.3}
\newcommand{\propdavge}{3.5}
\newcommand{\propfavge}{2.3}
\newcommand{\propgavge}{2.5}
\newcommand{\propiavge}{4.3}
\newcommand{\propsavge}{1.6}
\newcommand{\proptavge}{4.8}
\newcommand{\propwavge}{3.5}
\newcommand{\propaavgn}{3.9}
\newcommand{\propcavgn}{4.7}
\newcommand{\propdavgn}{4.4}
\newcommand{\propfavgn}{3.8}
\newcommand{\propgavgn}{3.8}
\newcommand{\propiavgn}{4.7}
\newcommand{\propsavgn}{3.4}
\newcommand{\proptavgn}{5.0}
\newcommand{\propwavgn}{4.3}
\newcommand{\propaavgp}{4.3}
\newcommand{\propcavgp}{2.0}
\newcommand{\propdavgp}{2.8}
\newcommand{\propfavgp}{4.4}
\newcommand{\propgavgp}{2.4}
\newcommand{\propiavgp}{2.3}
\newcommand{\propsavgp}{2.0}
\newcommand{\proptavgp}{2.0}
\newcommand{\propwavgp}{3.3}
\newcommand{\propaavgs}{3.1}
\newcommand{\propcavgs}{1.2}
\newcommand{\propdavgs}{1.9}
\newcommand{\propfavgs}{3.2}
\newcommand{\propgavgs}{2.0}
\newcommand{\propiavgs}{1.4}
\newcommand{\propsavgs}{1.9}
\newcommand{\proptavgs}{1.1}
\newcommand{\propwavgs}{2.3}

\begin{document}

\dochead{A Survey and Classification of Controlled Natural Languages}
\runningtitle{A Survey and Classification of Controlled Natural Languages}
\runningauthor{Tobias Kuhn}
\author{Tobias Kuhn\thanks{
Chair of Sociology, in particular of Modeling and Simulation, ETH Zurich, and
Institute of Computational Linguistics, University of Zurich.
E-mail:~\texttt{kuhntobias@gmail.com}.
Personal~website:~\texttt{http://www.tkuhn.ch}.
}}
\affil{
ETH Zurich and University of Zurich
}
\historydates{
Submission received: 26~October 2012;
revised version received: 7~March 2013;
accepted for publication: 25~April 2013.
}


\maketitle

\begin{abstract}
What is here called \textbf{controlled natural language (CNL)} has traditionally been given many different names. Especially during the last four decades, a wide variety of such languages have been designed. They are applied to improve communication among humans, to improve translation, or to provide natural and intuitive representations for formal notations. Despite the apparent differences, it seems sensible to put all these languages under the same umbrella. To bring order to the variety of languages, a general classification scheme is presented here. A comprehensive survey of existing English-based CNLs is given, listing and describing {\numberofcnls} languages from 1930 until today. Classification of these languages reveals that they form a single scattered cloud filling the conceptual space between natural languages such as English on the one end and formal languages such as propositional logic on the other. The goal of this article is to provide a common terminology and a common model for CNL, to contribute to the understanding of their general nature, to provide a starting point for researchers interested in the area, and to help developers to make design decisions.
\end{abstract}

\section{Introduction}

\emph{Controlled}, \emph{processable}, \emph{simplified}, \emph{technical}, \emph{structured}, and \emph{basic} are just a few examples of attributes given to constructed languages of the type to be discussed here. We will call them \textbf{controlled natural languages (CNL)} or simply \textbf{controlled languages}. Basic English, Caterpillar Fundamental English, SBVR Structured English, and Attempto Controlled English are some examples; many more will be presented below. This article investigates the nature of such languages, provides a general classification scheme, and explores existing approaches.

As the variety of attributes suggests, there is no general agreement on the characteristic properties of CNL, making it a very fuzzy term. There are two main reasons for this. First, CNL approaches emerged in different environments (industry, academia, and government), in different disciplines (computer science, philosophy, linguistics, and engineering), and over many decades (from the 1930s until today). People from different backgrounds often used and continue to use different names for the same kind of language. Second, although controlled natural languages seem to share important properties, they also exhibit a very wide variety: Some are inherently ambiguous, others are as precise as formal logic; virtually everything can be expressed in some, only very little in others; some look perfectly natural, others look more like programming languages; some are defined by just a handful of grammar rules, others are so complex that no complete grammar exists. This variety makes it difficult to get a clear picture of the fundamental properties.
This article aims at resolving this problem by giving an overview of existing CNLs and by providing a general classification scheme. Generally, this work has several, partly overlapping goals, ranging from purely theoretical to more practical objectives (listed in this order):
\begin{itemize}
\item To give us a better understanding of the nature of CNL
\item To establish a common terminology and a common model for CNL
\item To provide a starting point for researchers interested in CNL
\item To help CNL developers make design decisions
\end{itemize}

Although a wide variety of CNLs have been applied to a wide variety of problem domains, virtually all of them seem to be relevant to the field of computational linguistics. Among other techniques, they involve lexical analyses, grammar and style checking, ambiguity detection, machine translation, and computational semantics.

Unsurprisingly, most CNLs are based on English. For the sake of simplicity, the survey presented in this article is restricted to these languages and excludes existing approaches based on other natural languages, such as German and Chinese. The classification scheme to be presented, however, is general and not restricted to English in any way.

In what follows, the relevant background is discussed (Section~\ref{sec:background}), a classification scheme is introduced (Section~\ref{sec:pens}), existing English-based CNLs are classified and described based on a small sample (Section~\ref{sec:languages}), the results are analyzed (Section~\ref{sec:analysis}), and finally the conclusions are discussed (Section~\ref{sec:conclusions}). The appendix shows the full list of languages with short descriptions for each of them.

\section{Background}
\label{sec:background}

Controlled natural language being such a fuzzy term, it is important to clarify its meaning, to establish a common definition, and to understand the differences to related terms. In addition, it is helpful to review previous attempts to classify and characterize CNLs.

\subsection{Definition}
\label{sec:definition}

As mentioned above, there is no generally agreed-upon definition for controlled natural language and for closely related terms including controlled language, constrained natural language, simplified language, and controlled English. The following two quotations illustrate this:
\begin{quote}
A controlled language (CL) is a restricted version of a natural language which has been engineered to meet a special purpose, most often that of writing technical documentation for non-native speakers of the document language. A typical CL uses a well-defined subset of a language's grammar and lexicon, but adds the terminology needed in a technical domain.
\cite{kittredge2003oxfordcl}
\end{quote}
\begin{quote}
Controlled natural language is a subset of natural language that can be accurately and efficiently processed by a computer, but is expressive enough to allow natural usage by non-specialists.
\cite{fuchs1995clnlp}
\end{quote}
Both descriptions exhibit a strong bias towards one particular type of CNL (these types are discussed in more detail below): The first quotation focuses on technical languages that are designed to improve comprehensibility, whereas the second one only covers languages that can be interpreted by computers. They agree, however, on the fact that a CNL is based on a certain natural language but is more restrictive. It is also generally agreed that CNLs are constructed languages, which means languages that did not emerge naturally but have been engineered. The use of the term \emph{subset} is misleading though, since many CNLs are not proper subsets of the underlying natural language. Many of these languages have small deviations from natural grammar or semantics. Others make use of unnatural elements such as colors and parentheses to increase readability and precision. Some even consider the programming language COBOL a controlled natural language \cite{sowa2000ce}. The subset relation in its mathematical sense is clearly too strict to cover a large part of the languages commonly called CNL. Although they all clearly share important properties, the specific languages can be quite different in their coverage and nature. It is not surprising that O'Brian \shortcite{obrien2003eamtclaw}, who compared English-based CNLs of different types, comes to the conclusion that no common core language can be identified.
To meet these problems, the following definition is proposed here:
\begin{definition}[long]
A language is called a \textbf{controlled natural language} if and only if it has all of the following four properties:
\begin{enumerate}
\item It is based on exactly one natural language (its ``base language'').
\item The most important difference between it and its base language (but not necessarily the only one) is that it is more restrictive concerning lexicon, syntax, and/or semantics.
\item It preserves most of the natural properties of its base language, so that speakers of the base language can intuitively and correctly understand texts in the controlled natural language, at least to a substantial degree.
\item It is a constructed language, which means that it is explicitly and consciously defined, and \emph{is not} the product of an implicit and natural process (even though it is based on a natural language that \emph{is} the product of an implicit and natural process).
\end{enumerate}
\end{definition}
Properties 2 and 3 are deliberately vague, because it is not possible or desirable to draw a strict line there. Properties 1 and 3 refer to the N in CNL: naturalness; Properties 2 and 4 refer to the C: control. We will later be able to be a little more precise concerning property number 3. We leave it for now, and we can summarize this relatively verbose definition in the form of the following short version:
\begin{definition}[short]
A \textbf{controlled natural language} is a constructed language that is based on a certain natural language, being more restrictive concerning lexicon, syntax, and/or semantics while preserving most of its natural properties.
\end{definition}
As a further remark, we should note that the term \emph{language} is used in a sense that is restricted to sequential languages and excludes visual languages such as diagrams and the like. We can verify that the definitions above include virtually all languages that have been called CNL, while it excludes natural languages (since they are not constructed), languages such as Esperanto (since they are not based on one particular natural language), and common formal languages (since they lack intuitive understandability).

\subsection{Related Terms}

Before we move on to examine the types and properties of languages, we should discuss a number of terms that are related to CNL and are easy to confuse: sublanguage, fragments of language, style guide, phraseology, controlled vocabulary, and constructed language.

\textbf{Sublanguages} are languages that naturally arise when ``a community of speakers (i.e. `experts') shares some specialized knowledge about a restricted semantic domain [and] the experts communicate about the restricted domain in a recurrent situation, or set of highly similar situations'' \cite{kittredge2003oxfordcl}. As with controlled natural language, a sublanguage is based on exactly one natural language and is more restricted. The crucial difference between the two terms is that sublanguages emerge naturally, whereas CNLs are explicitly and consciously defined.

\textbf{Fragments of language} is a term denoting ``a collection of sentences forming a naturally delineated subset of [a natural] language'' \cite{pratthartmann2006ndjfl}. The term is closely related to CNL and the difference seems to be mainly a methodological one: Fragments of language are \emph{identified} rather than \emph{defined}, they are closely kept in the context of the full natural language and related fragments, and their purpose is rather to theoretically study them than to directly use them to solve a particular problem. A CNL can be seen as a fragment of a language ``developed for the purpose of supporting some technical activity'' \cite{pratthartmann2009cnl}.

\textbf{Style guides} are documents containing instructions on how to write in a certain natural language. Some style guides such as ``How to write clearly'' \cite{eu2011writeclearly} provide ``hints, not rules'' and therefore do not describe a new language, but only give advice on how to use the given natural language. However, other style guides such as the Plain Language Guidelines \cite{plain2011pl} are stricter and \emph{do} describe a language that is not identical to the respective full language. The question of whether such a language can be considered a CNL depends on whether the style guide defines a new language or whether it merely describes good practices that have emerged naturally.

\textbf{Phraseology} is a term that denotes a ``set of expressions used by a particular person or group'' \cite{houghton2000ahdel}. Typically, this term is used when the grammatical structure is simpler than in full natural language. In contrast to sublanguages and fragments of languages, a phraseology is not a selection of sentences but a selection of \emph{phrases}. Phraseologies can be natural or constructed, and in the latter case they are usually considered CNLs.

\textbf{Controlled vocabularies} are standardized collections of names and expressions, including ``lists of controlled terms, synonym rings, taxonomies, and thesauri'' \cite{ansiniso2005convoc}. Mostly, controlled vocabularies target a specific, narrow domain. In contrast to CNL, they do not deal with grammatical issues, that is, how to combine the terms to write complete sentences. Many CNL approaches, especially domain-specific ones, include controlled vocabularies.

\textbf{Constructed languages} (or \textbf{artificial languages} or \textbf{planned languages}) are languages that did not emerge naturally but have been consciously defined. In this broad sense, the term includes (but is not limited to) languages such as Esperanto, programming languages, and CNLs.

\subsection{Types and Properties}
\label{sec:types}

Let us now turn to the nature of CNLs. To bring order to their seemingly chaotic variety, more than 40 properties of such languages and their environments have been identified \cite{wyner2009cnlmain}. Many of these properties, however, are fuzzy and do not allow for a strict categorization. For the survey to be presented in Section \ref{sec:languages}, we collect nine general and clear-cut properties and give them letter codes. As it turns out, however, these properties mainly describe the application environment of languages and not so much the languages themselves. For that reason, a classification scheme is introduced in the next section to describe the fundamental nature of CNLs and other languages.

In general, controlled natural languages can be roughly subdivided according to the problem they are supposed to solve \cite{schwitter2002dexa}: to improve communication among humans, especially speakers with different native languages (we will use the letter code \prop{c} for these languages); to improve manual, computer-aided, semi-automatic, or automatic translation (\prop{t}); and to provide a natural and intuitive representation for formal notations (\prop{f}). The last type includes approaches for automatic execution of texts, which requires, at least conceptually, a mapping to an executable formalism.
As we will see, these three types emerged at different points in time: Type \prop{c} is the oldest, type \prop{t} emerged later, and type \prop{f} is the most recent of the three. Although this seems to be a sensible and useful subdivision, a simpler version based on just two types dominates the literature. Huijsen \shortcite{huijsen1998claw} introduced the distinction between ``human-oriented'' and ``computer-oriented'' languages. The former roughly corresponds to type \prop{c}, the latter to the types \prop{t} and \prop{f}. However, Huijsen observes that
``it is often difficult to qualify a controlled language as either human-oriented or machine-oriented, since often simplification works both ways.''
Because these types describe \emph{problems} rather than \emph{languages}, reusing a language in a different problem domain can change its type even if the language itself has not changed at all.
Other similar categorizations include the distinction between ``naturalistic'' (type \prop{c} and \prop{t}) and ``formalistic'' (type \prop{f}) languages \cite{pool2006claw,clark2009cnlmain} and the distinction between readability and translatability \cite{reuther2003eamtclaw}.

Another apparent fact is that some languages originated from academia (letter code \prop{a}), some from industry (\prop{i}), some from a government or a UN agency (\prop{g}), and others from a combination of the three. In addition, the distinction between general purpose languages and those for a particular restricted domain is often discussed \cite{pool2006claw}. This is related to the distinction of whether the lexicon is open or closed \cite{adriaens1992coling}. We will use the letter code \prop{d} to denote languages targeting a specific and narrow domain. A further important difference is the one between written and spoken languages. We will use \prop{w} to denote languages that are intended to be written, and \prop{s} for those that are intended to be spoken. However, none of these distinctions seems to describe a fundamental language property: Languages that originated in one environment can later be used in another; the lexicon can later be declared open or closed; written languages can be read aloud; and spoken languages can be written down.

The rules that define a CNL can be proscriptive or prescriptive \cite{nyberg2003translator}, or a combination of the two. Proscriptive rules describe what is \emph{not} allowed, whereas prescriptive rules describe what \emph{is} allowed. Languages defined by proscriptive rules alone must have some starting point in the form of a given (natural) language. Languages with only prescriptive rules, in contrast, typically start from scratch. As we will see, there is a close connection of this distinction to the concept of simplicity as introduced in the next section.

Because of their lack of generality, we do not include here more specific low-level properties such as the support for subclauses and free compounding \cite{adriaens1992coling}, specific restrictions on grammatical tenses and modal verbs \cite{obrien2003eamtclaw}, and support for interrogative and imperative sentences \cite{wyner2009cnlmain}.

Table \ref{tab:properties} summarizes the letter codes.
Any two of these properties can overlap, and therefore any combination is possible in theory (with the exception that no language should be neither \prop{w} nor \prop{s}).
\begin{table}[t]
\begin{center}
\caption{Letter codes for properties of CNLs}
\begin{tabular}{c|l}
Code & Property \\
\hline
\prop{c} & The goal is comprehensibility \\
\prop{t} & The goal is translation \\
\prop{f} & The goal is formal representation (including automatic execution) \\
\prop{w} & The language is intended to be written \\
\prop{s} & The language is intended to be spoken \\
\prop{d} & The language is designed for a specific narrow domain \\
\prop{a} & The language originated from academia \\
\prop{i} & The language originated from industry \\
\prop{g} & The language originated from a government \\
\end{tabular}
\label{tab:properties}
\end{center}
\end{table}

Finally, there is one additional aspect of constructed languages that deserves attention: their life cycle. Some languages are not much more than abstract ideas, others have left this stage being applied to concrete problems, and yet others have progressed to widespread application in productive environments. At different stages of maturity, languages can be discontinued or abandoned, which signifies the end of their life cycle. Obviously, these different stages flow into each other and it is often difficult to name a concrete year of birth or death (especially the latter, as most CNLs die silently). Where possible, we will keep track of these life cycle properties.

\section{PENS Classification Scheme}
\label{sec:pens}

As we have seen, the CNL properties introduced above describe application domains rather than the languages themselves. Certainly, several fundamental language properties have been identified and discussed in the literature, such as expressiveness \cite{mitamura1995tmi,boyd2005icre,pool2006claw}, complexity \cite{mitamura1995tmi}, grammar modifications \cite{pool2006claw}, understandability, natural look-and-feel, ambiguity, predictability, and formality of definition \cite{wyner2009cnlmain}. However, these properties are all very fuzzy and do not allow for strict categorization.

To construct a principled classification scheme for such fundamental language properties, it makes sense to condense them to a few dimensions that are to a large degree (though not entirely) independent of each other. Ambiguity, predictability, and formality of definition can be subsumed by a dimension that we can call \textbf{precision}. \textbf{Expressiveness} can make up a second dimension. Grammar modifications, understandability, and natural look-and-feel can be combined to a dimension of \textbf{naturalness}. A fourth dimension can be called complexity or --- to have a dimension of the type ``more is better'' --- \textbf{simplicity}.
This is how we arrive at the four dimensions Precision, Expressiveness, Naturalness, and Simplicity that underlie the \textbf{PENS} classification scheme.\footnote{These four dimension have first been sketched as ``design principles'' in the author's doctoral thesis \cite{kuhn2010doctoralthesis}, where ``precision'' was called ``clearness.''}

It seems that all fundamental language properties mentioned in the existing literature fall into one of these general dimensions, or can be broken down into different aspects that can be mapped to these dimensions. There are no strong dependencies between any two dimensions (for any dimension pair, it is easy to imagine languages that are at the top, bottom, and opposite ends in these two dimensions). Furthermore, there is no obvious dimension pair that could be merged in a meaningful way. Together, this seems to indicate that this set of dimensions is minimal yet complete.

The development of this scheme originated from the insight that CNLs can be conceptually located somewhere in the gray area between natural languages on the one end and formal languages on the other. Generally, CNLs are more formal than natural languages but more natural than formal ones. For instance, a natural language such as English is very expressive, but complex and imprecise. A formal language such as propositional logic, in contrast, is very simple and precise, but at the same time unnatural and inexpressive. CNLs must be somewhere in the middle, but where exactly?

It seems obvious that all four of the above-mentioned dimensions are continuous in nature or at least very fine-grained. In fact, one can argue that each of the dimensions is actually multidimensional and that representing it in one dimension is a rough simplification. Such simplifications are necessary, however, in order to get a precise measure for such vague concepts such as expressiveness.

Intuitively, PENS uses a natural language such as English and a formal language such as propositional logic as pegs to span a conceptual space in which different kinds of controlled natural languages can be placed. In order to get a general but strict classification scheme, PENS drastically simplifies things by restricting each of its four dimensions to five classes, to be numbered from 1 to 5.
These five classes are non-overlapping and consecutively cover the one-dimensional space between the two extremes: English on the one end and propositional logic on the other. For precision and simplicity, English is on the bottom end of the scale in class 1, which we write as P\textsuperscript{1} and S\textsuperscript{1}. Propositional logic is on the opposite end of the scale in class 5, represented with P\textsuperscript{5} and S\textsuperscript{5}. For expressiveness and naturalness, the roles are switched: English is at the top end (E\textsuperscript{5} and N\textsuperscript{5}) and propositional logic at the bottom (E\textsuperscript{1} and N\textsuperscript{1}). In this way, the scheme defines a conceptual space for CNLs that includes natural and formal languages as special cases.
Combining the four dimensions gives $5^4 = 625$ classes, represented with shorthand such as \pens1551 for English and \pens5115 for propositional logic. The difficult and interesting part of this intellectual exercise is where and how to draw the borders between the five classes of each dimension.

The decision to use five classes for each dimension, and not four or six, is somewhat arbitrary. A larger number of classes allows for more detailed classifications, whereas it also gets more difficult to come up with strict and objective criteria to define these classes. Five seems to be a good middle ground.

\subsection{Precision}

The precision dimension of the PENS scheme captures the degree to which the meaning of a text in a certain language can be directly retrieved from its textual form, that is, the sequence of language symbols. Natural language is very imprecise in this sense, because a large amount of context information is needed to grasp the meaning of typical sentences. Formal logic languages, on the other hand, have maximal precision, because their meaning is strictly defined solely on the basis of the possible sequences of their language symbols. The symbol grounding problem, that is, the problem of mapping symbols to their counterparts in the real world, is not considered here, because it affects all languages, including both natural and formal ones. On this precision dimension, languages are divided into the five classes P\textsuperscript{1}, P\textsuperscript{2}, P\textsuperscript{3}, P\textsuperscript{4}, and P\textsuperscript{5} as follows:

\paragraph{Imprecise languages (P\textsuperscript{1})} Virtually every sentence of these languages is vague to a certain degree. Without taking context into account, most sentences of a certain complexity are ambiguous. The automatic interpretation of such languages is ``AI-complete,'' which means it is a problem for which no complete solutions are in sight. These languages require a human reader to check whether a given statement is syntactically correct, and include borderline statements on which readers disagree. The same applies to the semantic properties of the language. All natural languages belong to this category.

\paragraph{Less imprecise languages (P\textsuperscript{2})} For these languages, the degree of ambiguity and vagueness is considerably lower than in natural languages, and their interpretation depends much less on context. They restrict the use and/or the meaning of a wide range of the respective ambiguous, vague, or context-dependent constructs. However, these constructs are still too dominant to make automatic interpretation reliable. Such languages are typically not related to a formal (i.e., mathematically precise) underpinning.

\paragraph{Reliably interpretable languages (P\textsuperscript{3})} The syntax of these languages is heavily restricted, though not necessarily formally defined. The restrictions are strong enough to make automatic interpretation reliable. There is a logical underpinning or at least a formal conceptual scheme, in which the semantics of sentences can be represented. However, the mapping of sentences to their formal representations is itself not defined in a fully formal way, but requires external background knowledge, heuristics, or user feedback.

\paragraph{Deterministically interpretable languages (P\textsuperscript{4})} Such languages are fully formal on the syntactic level; that is, they are (or can be) defined by a formal grammar. Each text in such a language can be deterministically parsed to a formal logic representation, or a small set of all possible representations (including all and only the possible ones). Based on the underlying formalism, these representations describe the meaning of the sentences, but they may be underspecified in the sense that they require certain parameters, background axioms, external resources, or heuristics to enable sensible deductions.

\paragraph{Languages with fixed semantics (P\textsuperscript{5})} These languages are fully formal and fully specified on both the syntactic and semantic levels. Each text has exactly one meaning, which can be automatically derived. The circumstances in which inferences hold or do not hold are fully defined. What conclusions follow from a given text in the language (e.g., whether it is consistent and which sentences of the language are a consequence of the text) can be defined with mathematical rigor, without the help of heuristics or external resources.

\subsection{Expressiveness}

The dimension of expressiveness describes the range of propositions that a certain language is able to express.
A language X is more expressive than a language Y if language X can describe everything that language Y can, but not vice versa. The relation of ``being more expressive'' does not constitute a total order: For two given languages of nonequal expressiveness, it can be the case (and often is the case) that neither is more expressive than the other. This entails that ranking a general set of languages in a linear order according to their expressiveness cannot be done in a completely objective way. A classification scheme, such as the one presented here, must therefore rely on only a subset of all possible expressiveness features. These expressiveness features should be general and important ones, and at the same time allow for a balanced and clear discrimination between the languages to be classified. The PENS classification scheme employs the following five expressiveness features:
\begin{enumerate}
\item[(a)] universal quantification over individuals (possibly limited)
\item[(b)] relations of arity greater than 1 (e.g., binary relations)
\item[(c)] general rule structures (\emph{if--then} statements with multiple universal quantification that can target all argument positions of relations)
\item[(d)] negation (strong negation or negation as failure)
\item[(e)] general second-order universal quantification over concepts and relations
\end{enumerate}
For each of these features to be considered fulfilled, they should be an integral part of the language and not just manifested by a few special cases.
There are a number of other important features that could be considered, for example support for existential quantification, equality, and types of supported speech acts (such as declarative, interrogative, directive, and indirect speech acts). However, to achieve a simple classification into a sequence of five classes, the features listed above will turn out to be sufficient and lead to a classification that seems consistent with the intuitive understanding of expressiveness.

Since this classification system should not only include declarative formal languages but also informal as well as procedural ones, it makes sense to apply a weaker notion of expressiveness than what is usually applied to logic languages. We can adopt from the research on programming languages the convention that a certain language construct adds expressiveness if its removal would require ``a global reorganization of the entire program'' \cite{felleisen1991scp}. If a certain language construct allows us to express something locally which would otherwise require us to reorganize the entire text, then we say that this language construct makes the language more expressive. This means, for example, that a language with second-order features relying on Henkin semantics qualifies for the last criterion of the above list, even though Henkin semantics can be reduced to first-order. A given set of statements written in a language with Henkin-style second-order features cannot generally be reduced to first-order logic without global reorganization, that is, changing statements that do not actually use second-order features. With this qualification, we can define the five classes as follows:

\paragraph{Inexpressive languages (E\textsuperscript{1})} These are languages lacking one or both of the features (a) and (b): They have no universal quantification or no relations of arity greater than 1. Propositional logic belongs to this category.

\paragraph{Languages with low expressiveness (E\textsuperscript{2})} Such languages have both of the features (a) and (b), but are not E\textsuperscript{3}-languages: They have universal quantification over individuals and relations of arity greater than 1. Description logics belong to this category.

\paragraph{Languages with medium expressiveness (E\textsuperscript{3})} These languages have all of the features (a), (b), (c), and (d), but are not E\textsuperscript{4}-languages: They have general rule structures and negation, in addition to the features of E\textsuperscript{2}. First-order logic belongs to this category.

\paragraph{Languages with high expressiveness (E\textsuperscript{4})} Such languages have all listed features (a), (b), (c), (d), and (e), but are not E\textsuperscript{5}-languages: They have second-order universal quantification over concepts and relations, in addition to the features of E\textsuperscript{3}. Second order predicate calculus belongs to this category.

\paragraph{Languages with maximal expressiveness (E\textsuperscript{5})} These languages can express anything that can be communicated between two human beings. Such languages cover any statement in any type of logic. Obviously, this includes all features listed above. All natural languages belong to this category.

\subsection{Naturalness}

The dimension of naturalness describes how close the language is to a natural language in terms of readability and understandability to speakers of the given natural language. We define the five classes as follows:


\paragraph{Unnatural languages (N\textsuperscript{1})} These are languages that do not look natural, making heavy use of symbol characters, brackets, or unnatural keywords. It might be possible to use natural words or phrases as names for certain entities, but this is neither required nor further defined by the language.

\paragraph{Languages with dominant unnatural elements (N\textsuperscript{2})} Natural language words or phrases are an integral part of such languages, but are dominated by unnatural elements or unnatural statement structure, or have unnatural semantics. The natural elements do not connect in a natural way to each other, and speakers of the given natural language typically fail to intuitively understand the respective statements.

\paragraph{Languages with dominant natural elements (N\textsuperscript{3})} In such languages, natural elements are dominant over unnatural ones and the general structure corresponds to natural language grammar. Due to the remaining unnatural elements or unnatural combination of elements, however, the sentences cannot be considered valid natural sentences. Speakers of the given natural language do not recognize the statements as well-formed sentences of their language, but are nevertheless able to intuitively understand them to a substantial degree.

\paragraph{Languages with natural sentences (N\textsuperscript{4})} These are languages with sentences that can be considered valid natural sentences. Speakers of the respective natural language recognize the statements as sentences of their language and are able to correctly understand their essence without instructions or training. Minor or infrequent exceptions and unnatural means for clarification (including text color, indentation, hyphenation, and capitalization) are permitted as long as they do not disturb the natural look-and-feel and the natural flow of the sentence. Parentheses and brackets in unnatural positions, however, in most cases \emph{do} disturb the natural text flow considerably, and are therefore typically not present in this category. While single sentences have a natural flow, this does not scale up to complete texts or documents. Complete texts in such languages seem very clumsy and repetitive, and lack a natural text flow.

\paragraph{Languages with natural texts (N\textsuperscript{5})} With these languages, complete texts and documents can be written in a natural style, with a natural text flow, and with natural semantics. In the case of spoken languages, complete dialogs can be produced with a natural flow and a natural combination of speech acts.

\bigskip\par

\noindent We can now be a little more precise concerning our definition of CNL. Property number 3 of the long version of the definition shown in Section \ref{sec:definition} says that a CNL ``preserves most of the natural properties of its base language, so that speakers of the base language can intuitively and correctly understand texts in the controlled natural language, at least to a substantial degree.'' We will interpret this in such a way that it only includes languages of naturalness N\textsuperscript{3} and higher. Thus, by this definition, there are no CNLs with N\textsuperscript{1} or N\textsuperscript{2}.

\subsection{Simplicity}

The fourth dimension is a measure of the simplicity or complexity of an exact and comprehensive language description covering syntax and semantics, if such a complete description is possible at all. This description should not presuppose intuitive knowledge about any natural language. It is therefore not primarily a measure for the effort needed by a human to learn the language, neither does it capture the theoretical complexity of the language (as, for example, the Chomsky hierarchy does). Rather, it is closely related to the effort needed to fully implement syntax and semantics of the language in a mathematical model, such as a computer program.

The PENS scheme applies a very pragmatic and simple indicator for simplicity: the number of pages in natural language needed to the describe the language in an exact and comprehensive way. For languages for which no such exact and comprehensive descriptions exist or can be written (that do not presuppose linguistic knowledge on the side of the reader, and given the current state of science), we can distinguish languages with the complexity of natural language from languages with considerably lower complexity.

These ``exact and comprehensive descriptions'' should define all syntactic and semantic properties of the language using accepted grammar notations to define the syntax and accepted mathematical or logical notations to define the semantics. They are assumed to use scientific writing style as found in scientific articles or technical reports, and should allow a skilled grammar engineer to implement a correct and complete parser within reasonable time. The page count should be based on a one-column format with up to about 700 words per page. It is important to note that the criterion is not the \emph{presence} of such a description but whether it is \emph{possible or not} to write one.

In order to treat languages with fixed vocabularies and those with extensible ones in an equal way, the above mentioned language descriptions do not need to include the vocabularies. Concretely, the five classes are defined as follows:

\paragraph{Very complex languages (S\textsuperscript{1})} These languages have the complexity of natural languages. They cannot be described in an exact and comprehensive manner.

\paragraph{Languages without exhaustive descriptions (S\textsuperscript{2})} These are languages that are considerably simpler than natural languages, in the sense that a significant part of the complex structures are eliminated or heavily restricted. Still, they are too complex to be described in an exact and comprehensive manner. Usually, the definitions of such languages just describe restrictions on top of a given natural language that is taken for granted.

\paragraph{Languages with lengthy descriptions (S\textsuperscript{3})} Such languages can be defined in an exact and comprehensive manner, but it requires more than ten pages to do so.

\paragraph{Languages with short descriptions (S\textsuperscript{4})} These are languages for which an exact and comprehensive description requires more than one page but not more than ten pages.

\paragraph{Languages with very short descriptions (S\textsuperscript{5})} Such very simple languages can be described in an exact and comprehensive manner on a single page.

\bigskip\par

\noindent S\textsuperscript{1} and S\textsuperscript{2} are considered complex because they rely on a given natural language. Coming back to a distinction briefly introduced in the previous section, such languages are typically defined by \emph{proscriptive} rules, describing what is not allowed compared to the full language. S\textsuperscript{3}, S\textsuperscript{4}, and S\textsuperscript{5}, in contrast, typically use \emph{prescriptive} rules that define the language from scratch. For that reason, they are simpler in our sense of the word than languages of the first type, which ``import'' the complexity of full natural language.

\bigskip\par

\noindent Before we move on to apply this scheme, it should be stressed that PENS is designed to measure the \emph{nature} of a language, not its \emph{quality} or \emph{usefulness}. It should be used to \emph{describe} languages, not to \emph{rank} them.
As the ``perfect'' language does not exist, compromises have to be made. Depending on application area, environment, and goal, different weights are assigned to the PENS dimensions, and therefore different optimal levels result.
In theory, more is better for each of the PENS dimensions, but this does not necessarily hold in practice. A certain level in any of the dimensions is often good enough for a given application domain, and going beyond that level brings no additional benefit.
Furthermore, as we restrict ourselves to just five classes per dimension, there can be relatively large differences \emph{within} one class. It is inevitable that two languages in the same class can be farther apart in the respective dimension than two languages in adjacent classes.
Even if a language has higher PENS values in every dimension than another language, this does not mean that the former is ``better'' in any meaningful sense of the word. Having a high PENS score for expressiveness, for example, just means that the general expressiveness level is high, and not that the language is able to express each and every statement of all languages with a lower score. Similarly, having a high score for naturalness does not mean that all aspects of the language are more natural as compared to all languages with a lower score.

\section{Languages}
\label{sec:languages}

We can now turn to the actual survey. For practical reasons, we restrict ourselves here to English-based languages, leaving out CNLs that are based on other languages, such as Chinese, French, German, Greek, Spanish, and Japanese \cite{pool2006claw}.
To give an overview of the different existing English-based CNLs, twelve important and influential languages are introduced below. The complete list can be found in the appendix; surprisingly, we ended up with exactly {\numberofcnls} languages. In addition, a handful of other languages for comparison are introduced below, such as natural English and propositional logic.
Each language is classified according to the nine properties with letter codes and the PENS scheme. A best guess is made in the cases where not enough information is available. The descriptions in the appendix are shorter in the case of similar languages or scarce information. This data set is also available online as a CSV table.\footnote{\url{http://purl.org/tkuhn/cnlsurvey/data}}

There are many user interface approaches based on some sort of natural language input, and it could be argued that they all --- at least indirectly --- define and use a controlled language, because none of them is able to correctly process full natural language. Such approaches, however, are included here only if the restrictions on the language are considered an inherent property of the approach and not a shortcoming of its implementation. In other words, the following listing excludes languages whose restrictions are not design decisions of the general approach but practical concessions, for example \namecite{warren1982ajcl}. The same criterion is applied to verbalization approaches, which inevitably define a restricted version of the respective language that could be considered a CNL, for example \namecite{halpin2004ista,jarrar2006verbalize}, and \namecite{lukichev2006i1d6}.
Other languages follow an approach called conceptual authoring or WYSIWYM \cite{hallett2007coli} where texts are created by short cycles of language generation and user-triggered modification actions. We include such languages here, because in this case the restrictions on the language are an important aspect of the approach. Finally, it should be mentioned that we leave out fictional languages, such as Newspeak of George Orwell's \emph{Nineteen Eighty-Four}.

Languages that do not have an official name are introduced by a ``generic name in quotation marks.'' Unless stated otherwise, quotes and examples are taken from the publications cited in the beginning of each paragraph.

\subsection{English-based Controlled Languages}
\label{sec:cnls}

Below, twelve selected CNLs are introduced, roughly in chronological order of their first appearance or the first appearance of similar predecessor languages. For this small sample, languages are chosen that were influential, are well-documented, and/or are sufficiently different from the other languages of the sample.

\makeatletter
\newenvironment{narrowquote}{\par\addvspace{4pt plus2pt}
      \extractfont\parindent18pt\noindent\ignorespaces
}{\par\ifx\@source\@empty\else{\sourcefont\noindent---\@source\par}\fi\gdef\@source{}\addvspace{3pt plus2pt}\@endparenv}
\makeatother

\newcommand{\langpara}[1]{\medskip\noindent\textbf{#1}}
\newcommand{\langparax}[3]{\medskip\noindent\textbf{#1}#3 --- #2}
\newcommand{\langexample}[1]{\begin{narrowquote}#1\end{narrowquote}}

\langparax{``Sowa's syllogisms''}{\pens5145, \prop{f~w~a}}{ \cite{sowa2000iccs} are simple logic languages based on the syllogisms originally introduced by \namecite{aristotle-350prioranalytics}. Sowa was probably the first to bring them into the context of CNL, claiming that they are the first reported instance of a controlled natural language. Because this survey is restricted to English, Sowa's version of the syllogisms is listed here instead of Aristotle's original version in ancient Greek. The complete language can be described by just four simple sentence patterns:
\begin{narrowquote}
Every A is a B. ~~~
Some A is a B. ~~~
No A is a B. ~~~
Some A is not a B.
\end{narrowquote}
A and B can be any English common nouns such as \emph{cat} and \emph{animal}. This language is very similar to the language $\mathcal{E}_0$ presented and studied by \namecite{pratthartmann2004jolli}, who used some additional patterns:
\begin{narrowquote}
Every A is not a B. ~~~
No A is not a B. ~~~
P is a B. ~~~
P is not a B.
\end{narrowquote}
Here, P can be any English proper name such as \emph{Socrates}.
We will use the term ``Sowa's syllogisms'' in a sense that includes such similar approaches.
The semantics of syllogisms is also very easy to define. The first four patterns shown above can be mapped to first-order logic like this (and similarly for the other patterns):
\begin{narrowquote}
$\forall x (~A(x) \rightarrow B(x)~)$ ~~~
$\exists x (~A(x) \wedge B(x)~)$ ~~~
$\neg \exists x (~A(x) \wedge B(x)~)$ ~~~
$\neg \forall x (~A(x) \rightarrow B(x)~)$
\end{narrowquote}
Hereby, we have an exact and comprehensive description of the language, taking just a couple of lines.
Despite the simple structure of the language, the sentences are perfectly natural. Its expressiveness, however, is very restricted: Only very simple sentence structures are covered and only one-place relations are supported.
}

\langparax{Basic English}{\pens2551, \prop{c~w}}{ \cite{ogden1930basic} is a language presented in 1930 that should improve communication among people around the globe. It is the first reported instance of a controlled version of English, at least the first one that received broader recognition. It influenced Caterpillar Fundamental English, which became itself a very influential language. Basic English was designed as a common basis for communication in politics, economy, and science.
It restricts the grammar and makes use of only 850 English root words. The restrictions are arguably most drastic in the case of verbs. Only 18 verbs are supported: put, take, give, get, come, go, make, keep, let, do, be, seem, have, may, will, say, see, and send. These verbs can be combined with prepositions to form more specific relations such as \emph{put in} to express \emph{insert}. Other verbs can be expressed with the help of nouns, such as \emph{give a move} instead of using \emph{move} as a verb. The usage of the given words and their variants is described by informal grammar rules, for example ``Collective nouns may be formed from adjectives when used with \emph{the}.'' These are two examples of sentences in Basic English:
\langexample{The camera man who made an attempt to take a moving picture of the society women, before they got their hats off, did not get off the ship till he was questioned by the police.}
\langexample{It was his view that in another hundred years Britain will be a second-rate power.}
Many variations exist that use larger word sets. The Simple English version of Wikipedia,\footnote{\url{http://simple.wikipedia.org}} for example, claims to use Basic English, but in fact uses a much less restricted language. Basic English is still used today and promoted by a dedicated Basic-English Institute.\footnote{\url{http://www.basic-english.org}} Many texts have been written in this language, including textbooks, novels, and large parts of the bible.
The drastic simplifications on the lexical level together with the grammatical restrictions constitute a significant gain in precision compared to full English. Still, any type of topic can be expressed with a natural text flow. The informal restrictions on the grammar, however, are not strong enough to significantly reduce the complexity of the language (in the PENS sense of complexity).
}

\langparax{E-Prime}{\pens1551, \prop{c~w~a}}{ or \textbf{E'} \cite{bourland1965gensem} is a restricted version of English with the only restriction being that the verb \emph{to be} is forbidden to use. This includes all inflectional forms such as \emph{are}, \emph{was} and \emph{being}, regardless of whether used as auxiliary or main verb. The language was presented in 1965 but the idea goes back to the late 1940s. The motivation for the use of E-Prime is the belief that ``dangers and inadequacies [...] can result from the careless, unthinking, automatic use of the verb `to be'.'' E-Prime is claimed by its proponents to enhance clarity.
The statement ``We do this because it is right'' would not be allowed, but one would have to rephrase it in a way that does not include \emph{to be}, for example:
\langexample{We do this thing because we sincerely desire to minimize the discrepancies between our actions and our stated ``ideals.''}
In the area of natural language processing, however, the verb \emph{to be} is \emph{not} considered one of the most difficult problems, which is good evidence that E-Prime is not considerably more precise than full English in the PENS sense.
Also in terms of complexity it is not considerably different from full English, because words such as \emph{become} and \emph{exist} are allowed that can replace the forbidden \emph{to be} in most cases. On the other hand, it seems true that it is always possible to rephrase a text without the use of \emph{to be} in a way that is fully natural though possibly longer than the original.
}

\langparax{Caterpillar Fundamental English (CFE)}{\pens2551, \prop{c~w~d~i}}{ \cite{verbeke1973traindev} was an influential controlled language developed at Caterpillar. It was officially introduced in 1971, was based on Basic English \cite{smart2003wft}, and has been reported to be the earliest industry-based CNL \cite{wojcik1997hlt}. The need for a controlled language emerged because of the increasing sophistication of Caterpillar's products and the need to communicate with non-English speaking service personnel in different countries \cite{verbeke1973traindev}: ``To summarize the problem: There are more than 20,000 publications that must be understood by thousands of people speaking more than 50 different languages.'' The idea of CFE was ``to eliminate the need to translate service manuals'' \cite{kamprath1998claw}. A trained, non-English speaking mechanic familiar with Caterpillar's products should be able to understand the language after completing a course on CFE consisting of 30 lessons. The vocabulary of the language is restricted to around 800 to 1,000 words \cite{crabbe2009ptc}, with only one meaning defined for each of them (e.g., \emph{right} only as the opposite of \emph{left}). Still, many of the words ``had broad semantic scope and it was assumed that they would be disambiguated in context by the human reader'' \cite{kamprath1998claw}. The following ten rules summarize the grammatical restrictions \cite{crabbe2009ptc}:
\begin{center}\small
\begin{minipage}[t]{0.475\textwidth}
1. Make positive statements.\\
2. Avoid long and complicated sentences.\\
3. Avoid too many subjects in one sentence.\\
4. Avoid too many successive adjectives and\\
\phantom{4.} nouns.\\
5. Use uniform sentence structures.
\end{minipage}
\begin{minipage}[t]{0.475\textwidth}
6. Avoid complicated past and future tenses.\\
7. Avoid conditional tenses.\\
8. Avoid abbreviations, contractions, and\\
\phantom{8.} colloquialisms.\\
9. Use punctuation correctly.\\
10. Use consistent nomenclature.
\end{minipage}
\end{center}
These are two examples of CFE sentences:
\langexample{The maximum endplay is .005 inch.}
\langexample{Lift heavy objects with a lifting beam only.}
CFE was discontinued by Caterpillar in 1982, because (among other reasons) ``the basic guidelines of CFE were not enforceable in the English documents produced'' \cite{kamprath1998claw}. As a result, Caterpillar Technical English (see appendix) was developed following a different approach: The restrictions on the language should be enforceable, and should reduce translation costs instead of trying to eliminate the need for translations altogether.
The strong lexical restrictions together with some grammatical constraints make CFE more precise than full English, but it is not considerably different in terms of expressiveness, naturalness, and complexity.
}

\langparax{FAA Air Traffic Control Phraseology}{\pens2132, \prop{c~s~d~g}}{ \cite{faa2010atc}
is a controlled language defined by the Federal Aviation Administration (FAA) and used for the communication in air traffic coordination, going back to at least the early 1980s. There are other very similar languages for air traffic control such as the ICAO and CAA phraseologies. To a large extent, these languages are indistinguishable from each other, and together they are sometimes called \textbf{AirSpeak} \cite{robertson1987airspeak}. The FAA Phraseology is defined by more than 300 fixed sentence patterns such as ``(ACID), IN THE EVENT OF MISSED APPROACH (issue traffic). TAXIING AIRCRAFT/VEHICLE LEFT/RIGHT OF RUNWAY.'' This is an example of a statement following that pattern:
\langexample{United 623, in the event of missed approach, taxiing aircraft right of runway.}
In addition to these explicit patterns, there are many more implicit patterns defined in prose form, for example ``Issue advisory information on [...] bird activity. Include position, species or size of birds, if known, course of flight, and altitude.'' The following statement is an example that corresponds to this implicit pattern:
\langexample{Flock of geese, one o'clock, seven miles, northbound, last reported at four thousand.}
Vocabulary and semantics are restricted too, for example ``Use the word \emph{gain} and/or \emph{loss} when describing to pilots the effects of wind shear on airspeed.'' Phraseology statements can be mixed with statements in full English in cases where no pattern exists to express the desired message.
The language is heavily restricted and much less ambiguous than full English.
It is inexpressive in the sense that no universal quantification is supported, and is not sufficiently restricted to make an exact and exhaustive description feasible.
}

\langparax{ASD Simplified Technical English (ASD-STE)}{\pens2551, \prop{c~t~w~d~i}}{ \cite{asd2013ste}, often abbreviated to \textbf{Simplified Technical English (STE)} or just \textbf{Simplified English},
 is a CNL for the aerospace industry. Originally inspired by a language called ILSAM \cite{adriaens1992coling}, the language had its origins in 1979, but it was only in 1986 when it was officially presented for the first time, then under the name \textbf{AECMA Simplified English}. It received its current name in 2004 when AECMA merged with two other associations to form ASD.
The main purpose of the language is to make texts easier to understand, especially for non-native speakers. While AECMA Simplified English was designed to make translation into other languages unnecessary, one of the original goals of ASD-STE was to improve translation.
Today, the language is maintained by the Simplified Technical English Maintenance Group.
ASD-STE is based on English with restrictions expressed in about 60 general rules. These rules restrict the language on the lexical level (e.g., ``Use approved words from the Dictionary only as the part of speech given''), on the syntactic level (e.g., ``Do not make noun clusters of more than three nouns''), as well as on the semantic level (e.g., ``Keep to the approved meaning of a word in the Dictionary. Do not use the word with any other meaning.''). There is a fixed vocabulary consisting of terms common to the aerospace domain. Additionally, user-defined ``Technical Names'' and ``Technical Verbs'' can be introduced. This is an exemplary excerpt of a text in ASD-STE:
\langexample{These safety precautions are the minimum necessary for work in a fuel tank. But the local regulations can make other safety precautions necessary.}
Even though its restrictions make ASD-STE considerably more precise than full English, it does not allow for reliable automatic interpretation.
Full expressiveness and full naturalness of unconstrained English are retained, but also its complexity.
}

\langparax{Standard Language (SLANG)}{\pens3142, \prop{c~f~w~d~i}}{ \cite{rychtyckyj2002amta,rychtyckyj2005ai} is a language developed at Ford Motor Company starting from 1990. It is designed for process sheets containing build instructions for component and vehicle assembly plants. It is still used at Ford and has been continually extended and updated to reflect technical and business-related advances. With SLANG, engineers can write instructions that are clear and concise and at the same time machine-readable. Based on these instructions, the system can, among other things, automatically generate a list of required elements and calculate labor times. In addition, the restricted nature of the language is exploited to translate such instructions with the help of machine translation for their use in assembly plants in different countries. All SLANG sentences are in imperative mood and follow a certain general pattern starting with a main verb and followed by a noun phrase. There are additional restrictions on vocabulary and semantics. These are two exemplary sentences:
\langexample{OBTAIN ENGINE BLOCK HEATER ASSEMBLY FROM STOCK}
\langexample{APPLY GREASE TO RUBBER O-RING AND CORE OPENING}
A parser is used to check for compliance with the restrictions.
English grammar is followed with some minor deviations: For example, articles can be dropped and some kinds of modifiers can be used in unnatural ways.
}

\langparax{SBVR Structured English}{\pens3442, \prop{c~f~w~i}}{ \cite{omg2008sbvr} is a CNL for business rules first presented around 2005. It is part of the Semantics of Business Vocabulary and Business Rules (SBVR) standard. It was probably influenced by a language called RuleSpeak that is very similar and was first presented in 1994. The vocabulary is extensible and consists of four types of sentence constituents: terms (i.e., concepts), names (i.e., individuals), verbs (i.e., relations), and keywords (e.g., fixed phrases, quantifiers, and determiners). Each of these has its own color and style, as the following examples show:
\newcommand{\sbvrt}[1]{{\sffamily\color{Emerald}\uline{#1}}}
\newcommand{\sbvrn}[1]{{\sffamily\bfseries\color{Emerald}\uuline{#1}}}
\newcommand{\sbvrv}[1]{{\sffamily\itshape\color{blue}#1}}
\newcommand{\sbvrk}[1]{{\sffamily\color{Orange}#1}}
\langexample{\sbvrk{A} \sbvrt{rental} \sbvrk{must} \sbvrv{be guaranteed by} \sbvrk{a} \sbvrt{credit card} \sbvrv{before} \sbvrk{a} \sbvrt{car} \sbvrv{is assigned to} \sbvrk{the} \sbvrt{rental}\sbvrk{.}}
\langexample{\sbvrn{Rentals by Booking Mode} \sbvrv{contains} \sbvrk{the} \sbvrt{categories} \sbvrk{'}\sbvrt{advance rental}\sbvrk{'} \sbvrk{and} \sbvrk{'}\sbvrt{walk-in rental}\sbvrk{.'}}
The SBVR standard provides formal semantics based on second-order logic with Henkin semantics. The second of the above examples makes use of the second-order features. Some keywords have a precise meaning, such as \emph{or} meaning inclusive logical disjunction (unless followed by \emph{but not both}). Other keywords, however, are less precise, such as the determiner \emph{a} being defined as ``universal or existential quantification, depending on context based on English rules.'' The language strictly defines the permissible sentence constituents, but is much less strict in defining the order in which these constituents can be put. The syntax structure can be ambiguous (e.g., when using \emph{and} and \emph{or} in the same sentence), and so can be quantifier scopes and anaphoric references. There is no formal grammar of the language, and its definition depends to some degree on the linguistic understanding of a human reader.
}

\langparax{Attempto Controlled English (ACE)}{\pens4343, \prop{f~w~a}}{ \cite{fuchs2008reasoningweb} is a CNL with an automatic and unambiguous translation into first-order logic. ACE was first presented in 1996 as a language for software specifications. Later, the focus shifted towards knowledge representation and the Semantic Web. The language has been extended over the years in various ways. The most notable features of ACE include complex noun phrases, plurals, anaphoric references, subordinated clauses, modality, and questions. These are two exemplary ACE sentences:
\langexample{A customer owns a card that is invalid or that is damaged.}
\langexample{Every continent that is not Antarctica contains at least 2 countries.}
ACE sentences are deterministically mapped to discourse representation structures, a notational variant of first-order logic. These expressions, however, are underspecified in the sense that many deductions (e.g., when involving plurals or modal verbs) require external background axioms that are not fixed by the ACE definition (these axioms are external in the sense that they are not necessarily expressible in ACE). This makes it possible to use ACE in different areas such as ontology editors, rule systems, and general reasoners with semantics that are not fully compatible. ACE is, with a few minor exceptions, fully natural on the sentence level, but longer texts do not have a natural text flow. Recently, ACE has also been used in the context of rule-based machine translation \cite{kaljurand2013eswc}, but translation was not a stated goal during the design of the language.}

\langparax{``Drafter Language''}{\pens4143, \prop{f~w~d~a}}{ \cite{power1998acl} is a CNL used in a system called Drafter-II. The language is used for instructions to word processors and diary managers.
The system employs a conceptual authoring approach: Users cannot directly edit the CNL text, but they can only trigger modification actions starting from a small stub sentence. In this way, incomplete statements are gradually completed by the user. The following example is a sequence of two incomplete statements showing one such completion step:
\langexample{Schedule \textbf{this event} by applying \emph{this method}.}
\langexample{Schedule the appointment by applying \emph{this method}.}
The first sentence has two missing parts: \emph{this event} and \emph{this method}. At this point, the user can choose, for example, \emph{the appointment} to fill in the first missing part, which leads to the second statement, which is still incomplete but has only one missing part left.
Once a statement is completed, Drafter-II internally maps it to Prolog expressions, which are then automatically executed.
As structural ambiguity can be resolved based on the given sequence of modification actions, languages following the conceptual authoring approach do typically not attempt to fully eliminate structural ambiguity. A given text can have multiple parse trees, only one of which corresponds to the way it was created.}

\langparax{E2V}{\pens5244, \prop{f~w~a}}{ \cite{pratthartmann2003jolli} is a controlled language that was introduced in 2001 and corresponds to the language $\mathcal{E}_3$ studied in later work \cite{pratthartmann2004jolli}. The ultimate goal is ``to provide useable tools for natural language system specification.'' E2V deterministically maps to the two-variable fragment of first-order logic. Because of this, satisfiability of E2V sentences and texts is decidable and computation is NEXPTIME complete. Two examples of E2V sentences are shown here:
\langexample{Some artist does not despise every beekeeper.}
\langexample{Every artist who employs a carpenter despises every beekeeper who admires him.}
The language is defined by 15 simple grammar rules plus nine predefined lexical rules for general words such as \emph{every} and \emph{does not}. A separate, user-defined lexicon contains the domain-specific words such as \emph{artist} and \emph{admires}.
Altogether, E2V is a precise, natural, simple, but relatively inexpressive controlled language.
}

\langparax{Formalized-English (FE)}{\pens5433, \prop{f~w~a}}{ \cite{martin2002iccs} is a CNL for knowledge representation. It is based on Conceptual Graphs and the Knowledge Interchange Format (KIF), and focuses on expressiveness. It covers a wide range of features including general universal quantification, negation, contexts (statements about statements), lambda abstractions, possibility, collections, intervals, and higher-order statements (reducible to first-order logic). Two examples of statements in FE are shown here (the second one is higher-order):
\langexample{At least 93\% of [bird with chrc a good health] can be agent of a flight.}
\langexample{If `a binaryRelationType *rt has for chrc the transitivity' then `if `\^{}x has for *rt \^{}y that has for *rt \^{}z' then `\^{}x has for *rt \^{}z' '.}
FE looks natural for simple statements, but becomes quite unnatural for more complex ones. This is due to unnatural use of parentheses, quotation marks, variables, and keywords such as \emph{chrc}. The syntax of the language is defined by about 50 rules in a parser generator language.}

\subsection{Languages for comparison}

For the analysis to be described in the next section, we will use the following languages for comparison, which are \emph{not} CNLs according to our definition:


\langparax{English}{\pens1551, \prop{c~w~s}}{ is our representative of a natural language.}

\langparax{Propositional logic}{\pens5115, \prop{f~w~a}}{ is a very basic logic language.}

\langparax{First-order logic}{\pens5314, \prop{f~w~a}}{ can be considered an extension of propositional logic. It is more expressive, but also more complex.}

\langparax{COBOL}{\pens5223, \prop{f~w~a~i~g}}{ is one of the oldest programming languages, which some call a controlled natural language \cite{sowa2000ce}. This is an exemplary COBOL statement:
\langexample{
PERFORM P WITH TEST BEFORE VARYING C FROM 1 BY 2 UNTIL C GREATER THAN 10.
}
Although COBOL uses natural phrases where other programming languages use symbols or short keywords, the statement structure does not really follow natural grammar. For that reason, we do not consider it a CNL.}


\langparax{Manchester OWL syntax}{\pens5224, \prop{f~w~a}}{ \cite{horridge2006owled} is a user-friendly syntax for the ontology language OWL. This is an exemplary expression:
\langexample{Pizza and not (hasTopping some FishTopping) and not (hasTopping some MeatTopping)}
Instead of logical symbols, natural words such as \emph{not} and \emph{some} are used. The general structure, however, resembles formal and not natural languages, which is why we do not consider it a CNL.
}

\bigskip

\noindent Naturally, there are many more languages that could be used for comparison, but the list above seems to be a good sample.

\section{Analysis}
\label{sec:analysis}

The data presented in the previous section and in the appendix allow for different kinds of aggregations and analyses. In particular, the classes and properties of the observed languages and the timeline of their evolution are interesting.

\subsection{PENS Classes}

Table \ref{tab:classes} summarizes the PENS classes and properties of the discussed CNLs. Some interesting patterns can be found in these data. Theoretically, there are $5^4 = 625$ possible PENS classes, but not all of them are observed ``in the wild.'' Some are even practically impossible, as far as we can tell, such as the perfect class \pens5555. The collected CNLs cover 25 distinct classes, which might seem a small number with respect to the entire PENS space, but they are, as we will see, widely scattered. Even though some hotspots of classes and properties can be identified, the languages exhibit a broad variety.

\begin{table}[tp]
\begin{center}
\caption{Observed PENS classes and properties of CNLs (sorted by PENS class)}
\scriptsize
\begin{tabular}{@{}l@{~~}l@{~~}p{10.9cm}@{}}
class & properties & languages
 \\
\hline
\pens1551 & \prop{c t w i} & IBM's EasyEnglish \\
 & \prop{c w s g} & Special English \\
 & \prop{c w a} & E-Prime \\
 & \prop{c w g} & Plain Language \\
\pens2132 & \prop{c s d g} & CAA Phraseology, FAA Phraseology, ICAO Phraseology, PoliceSpeak, SEASPEAK \\
\pens2133 & \prop{c w d i} & Airbus Warning Language \\
\pens2541 & \prop{f w a} & AIDA \\
\pens2551 & \prop{c t w d a i} & ALCOGRAM, COGRAM \\
 & \prop{c t w d a} & CLCM \\
 & \prop{c t w d i} & ASD-STE, Avaya CE, Bull GE, CTE, CASL, CE at Douglas, DCE, General Motors GE, PACE, Sun Proof \\
 & \prop{c t w d} & Wycliffe Associates' EasyEnglish \\
 & \prop{c t w i} & iCE, SMART Controlled English \\
 & \prop{c w d i} & AECMA-SE, CFE, CASE, CE at Clark, CE at IBM, CE at Rockwell, EE, HELP, ILSAM, KISL, NCR FE \\
 & \prop{c w d g} & Massachusetts Legislative Drafting Language \\
 & \prop{c w i} & Boeing Technical English, NSE, SMART Plain English \\
 & \prop{c w} & Basic English \\
 & \prop{t w d i} & MCE, Oc\'e Controlled English \\
 & \prop{t w a} & KCE \\
 & \prop{t w i} & CLOUT \\
\pens3142 & \prop{c f w d i} & SLANG \\
 & \prop{f s d i} & Voice Actions \\
\pens3243 & \prop{f w d a} & RNLS \\
\pens3333 & \prop{f w a} & ClearTalk \\
 & \prop{f w i} & ITA CE \\
\pens3342 & \prop{f w i} & CPL \\
\pens3442 & \prop{c f w i} & RuleSpeak, SBVR-SE \\
\pens4143 & \prop{f w d a} & Drafter Language, MILE Query Language \\
\pens4144 & \prop{f w a} & Quelo Controlled English \\
\pens4153 & \prop{t f d a} & PILLS Language \\
\pens4243 & \prop{f w d a} & Atomate Language \\
 & \prop{f w a i} & Gellish English \\
 & \prop{f w a} & GINO's Guided English \\
 & \prop{f w i} & CELT \\
\pens4343 & \prop{f w d a} & PROSPER CE \\
 & \prop{f w a} & ACE \\
\pens4353 & \prop{f w d a} & ICONOCLAST Language \\
\pens5143 & \prop{f w d a} & CLEF Query Language \\
 & \prop{f w a} & Ginseng's Guided English \\
\pens5144 & \prop{f w d a} & Coral's Controlled English \\
 & \prop{f w a} & PathOnt CNL \\
\pens5145 & \prop{f w a} & Sowa's syllogisms \\
\pens5234 & \prop{f w d a i} & TBNLS \\
 & \prop{f w a} & OWLPath's Guided English, SQUALL \\
\pens5243 & \prop{f w a} & CPE, CLIP, OWL ACE, SOS \\
\pens5244 & \prop{f w d a} & BioQuery-CNL, PERMIS CNL, ucsCNL \\
 & \prop{f w a} & CLOnE, DL-English, E2V, Lite Natural Language, OSE \\
 & \prop{f w g} & Rabbit \\
\pens5333 & \prop{f w d a} & CLM, ForTheL, Naproche CNL \\
 & \prop{f w a} & CLCE, PNL \\
\pens5343 & \prop{f w d a} & Gherkin \\
 & \prop{f w a g} & RECON \\
 & \prop{f w a} & First Order English, PENG, PENG-D, PENG Light \\
 & \prop{f w i} & iLastic Controlled English \\
\pens5433 & \prop{f w a} & FE
\end{tabular}

\label{tab:classes}
\end{center}
\end{table}

Visualization of the languages in the conceptual space can give us a better picture of the data. Since the PENS scheme is four-dimensional, it is difficult to visualize all dimensions in a single diagram. Figure \ref{fig:pens} shows a diagram for each of the six possible dimension pairs: The dots represent CNLs in comparison to natural languages such as English (white dot) and common formal languages (black dots). Note that the dots represent PENS classes and not individual languages.

\begin{figure}[tp]
\begin{center}%
\hspace{-0.4mm}%
\scalebox{0.7}{\input{diagram_pe}}%
\hfill%
\scalebox{0.7}{\input{diagram_pn}}%
\hfill%
\scalebox{0.7}{\input{diagram_ps}}%
\medskip\\%
\hspace{-0.4mm}%
\scalebox{0.7}{\input{diagram_ns}}%
\hfill%
\scalebox{0.7}{\input{diagram_es}}%
\hfill%
\scalebox{0.7}{\input{diagram_en}}%
\end{center}
\caption{Visualization of the PENS dimensions of existing CNLs, as compared to natural languages (white dot) and common formal languages (black dots). Each dot represents a PENS class containing one or more languages.}
\label{fig:pens}
\end{figure}

It is evident that the CNLs are widely scattered between the two extreme cases of natural English (white dot) and propositional logic (black dot in the corner). Seen from any angle, the set of existing CNLs exhibits wide variation. Except for the subspace with a naturalness level of less than 3, where there can be no CNLs by our definition, they cover a large part of the conceptual space. This indicates that PENS is a powerful scheme for distinguishing different CNLs.

The diagrams also show that the CNL classes form one single cloud, from any perspective, and not two or more disconnected clouds. This means that it would be difficult to come up with a clean categorization scheme that would subdivide the large and diverse set of existing CNLs. This seems to justify the decision of using the term CNL in a broad sense and not replacing it by more specific terms.

For several dimension pairs, strong correlations are observed. Precision and simplicity are positively correlated: More precise languages tend to be simpler (Spearman's rank correlation coefficient $\rho = 0.90$, using individual languages as data points and excluding the languages for comparison). Expressiveness and simplicity exhibit a strong negative correlation: More expressive languages tend to be more complex ($\rho = -0.82$). In addition, naturalness/expressiveness are strongly positively ($\rho = 0.77$) and naturalness/simplicity strongly negatively correlated ($\rho = -0.76$). At a slightly lesser degree, negative correlation values are obtained for the pairs precision/naturalness ($\rho = -0.67$) and precision/expressiveness ($\rho = -0.66$). These observations seem to be in line with what one would intuitively expect.

\subsection{Properties}

\begin{table}[tp]
\begin{center}
\caption{Properties of existing CNLs with average PENS values}
\small
\begin{tabular}{c@{~~~}l|r|r@{~~~}r@{~~~}r@{~~~}r@{~~~}r@{~~~}r@{~~~}r@{~~~}r@{~~~}r|l@{~~~~}l@{~~~~}l@{~~~~}l}
& & total & \multicolumn{9}{c|}{combined with property} & \multicolumn{4}{c}{PENS average} \\
& property & & \prop{c} & \prop{t} & \prop{f} & \prop{w} & \prop{s} & \prop{d} & \prop{a} & \prop{i} & \prop{g} & P & E & N & S \\
\hline
\prop{c} & comprehensibility      & \propcabs
  & \phantom{0}- & \propct & \propcf & \propcw & \propcs & \propcd & \propca & \propci & \propcg
  & \propcavgp & \propcavge & \propcavgn & \propcavgs \\
\prop{t} & translation            & \proptabs
  & \propct & \phantom{0}- & \proptf & \proptw & \propts & \proptd & \propta & \propti & \proptg
  & \proptavgp & \proptavge & \proptavgn & \proptavgs \\
\prop{f} & formal representation  & \propfabs
  & \propcf & \proptf & \phantom{0}- & \propfw & \propfs & \propfd & \propfa & \propfi & \propfg
  & \propfavgp & \propfavge & \propfavgn & \propfavgs \\
\prop{w} & written                & \propwabs
  & \propcw & \proptw & \propfw & \phantom{0}- & \propws & \propwd & \propwa & \propwi & \propwg
  & \propwavgp & \propwavge & \propwavgn & \propwavgs \\
\prop{s} & spoken                 & \propsabs
  & \propcs & \propts & \propfs & \propws & \phantom{0}- & \propsd & \propsa & \propsi & \propsg
  & \propsavgp & \propsavge & \propsavgn & \propsavgs \\
\prop{d} & domain-specific        & \propdabs
  & \propcd & \proptd & \propfd & \propwd & \propsd & \phantom{0}- & \propda & \propdi & \propdg
  & \propdavgp & \propdavge & \propdavgn & \propdavgs \\
\prop{a} & academia               & \propaabs
  & \propca & \propta & \propfa & \propwa & \propsa & \propda & \phantom{0}- & \propai & \propag
  & \propaavgp & \propaavge & \propaavgn & \propaavgs \\
\prop{i} & industry               & \propiabs
  & \propci & \propti & \propfi & \propwi & \propsi & \propdi & \propai & \phantom{0}- & \propig
  & \propiavgp & \propiavge & \propiavgn & \propiavgs \\
\prop{g} & government             & \propgabs
  & \propcg & \proptg & \propfg & \propwg & \propsg & \propdg & \propag & \propig & \phantom{0}-
  & \propgavgp & \propgavge & \propgavgn & \propgavgs \\
\end{tabular}
\label{tab:props}
\end{center}
\end{table}

Let us turn to the properties. Table \ref{tab:props} shows the number of CNLs for each of the properties we considered and their combinations. As some languages have been used more extensively and over longer periods of time than others, these numbers do not necessarily reflect the actual importance or popularity of the different language types. The table also shows the average PENS values for each type. Again, we should be careful when interpreting these numbers, as all languages were equally weighted, which does not take into consideration that some languages are much more mature and wide-spread than others. Nevertheless, these numbers reveal some interesting facts.

For a bit less than half of the languages, the goal is to increase comprehensibility. Formal representation is the goal of another, only slightly overlapping, half. About 22\% of all languages have translatability as their goal. There is a large overlap of the types \prop{c} and \prop{t}, whereas these two barely overlap with \prop{f}. Existing CNL approaches can therefore be roughly subdivided into two groups of similar size: one consisting of languages for improved comprehensibility and translatability, and the other made up of languages that have formal representation as their goal. Mostly, languages of the types \prop{c} and \prop{t} are domain-specific, originated from industry, and  focus more on expressiveness and naturalness than on precision or simplicity. Languages of type \prop{f}, in contrast, mostly have an academic origin and tend to have a much stronger focus on precision and simplicity at the cost of expressiveness and naturalness.

When it comes to the distinction between written and spoken languages, we see a very one-sided picture: More than 90\% of all languages are intended to be written; we found only seven languages that are intended to be spoken (one of which is intended to be spoken \emph{and} written). The reason for this might be that controlling a spoken language is much more difficult in practice. Written texts can be revised and given to a language checker before publication, whereas spoken language typically lacks this two-stage process. It is an interesting fact that six out of the seven spoken languages originated from a governmental environment. On average, written languages have higher PENS values in all four dimensions.

Concerning domain-specificity, the data are balanced. About half of the languages are designed for a specific and narrow domain. The other half follow a more general-purpose approach. Comprehensibility is the prevalent goal for domain-specific languages, and they mostly originated from industry. No clear tendencies can be identified with respect to the PENS dimensions.

Concerning the last three properties, the data show similar language counts for academic and industrial CNLs: {\propaabs} and {\propiabs} languages, respectively. On the other hand, only ten CNLs were found that originated from a governmental environment. It has to be noted, however, that information about CNLs from industry is typically much scarcer than about languages from academia or governments. It is therefore likely that most of the languages that escaped this survey (because of missing or hard-to-find information) are industrial ones. Such a bias might also be present in the case of some of the other properties discussed above. In any case, academia apparently focuses much more on languages for formal representation than for comprehensibility or translation, whereas industry seems to have an opposite focus.

\subsection{Design Decisions}

Apart from being a description of the current state of the art, Table \ref{tab:props} can be a valuable tool for making design decisions when creating a new CNL. In such a situation, the application environment of the language to be defined is typically fixed, but not yet the inherent properties of the language itself. Those inherent language properties are supposed to be fixed only during the design process. At the early design stage, the table above can be used to check the level of previous work on CNLs for a given combination of environment properties. It also delivers the PENS classes of a typical CNL in this environment, which can be used to guide the design process.

For example, if you intend to create a domain-specific, industrial CNL to enhance comprehensibility, the table tells you that the combination of these properties is not unusual at all (at least pair-wise combinations). Furthermore, the table indicates that such a language typically has a PENS class somewhere between \pens2341 and \pens3552.
As a second example, somebody might want to design a CNL for speech translation. A quick look at the table reveals that no such CNL has been reported so far, which indicates that a significant amount of original work is needed for the design of such a language. We also see that a typical spoken CNL is very different from a typical language for translation in terms of expressiveness and naturalness. This suggests two important design decisions: How expressive should the resulting language be, and how natural?

The table can reveal such questions about design decisions, but of course it cannot answer them. Nevertheless, such information about existing approaches in similar problem domains and environments can be very valuable to focus the design effort to the crucial aspects.

\subsection{Timeline}

Since CNLs have been defined and used over many decades and have influenced each other, it is interesting to draw the evolution of these languages on a timeline, as Figure~\ref{fig:timeline} does. Each bar represents the ``life'' of a language, that is, the period when the language was studied or used. For some languages, the year of ``birth'' or ``death'' is unknown, which is indicated by dashed bars fading in and out. The vertical lines show influences from other languages at the time of birth (solid for reported influences; dashed for influences that are not reported but seem probable). The colors of the bars represent the goals of the languages, as indicated in the legend.

\definecolor{propc} {rgb}{1.00,1.00,1.00}
\definecolor{propct}{rgb}{0.65,0.65,1.00}
\definecolor{propt} {rgb}{0.30,0.30,1.00}
\definecolor{proptf}{rgb}{0.25,0.25,0.60}
\definecolor{propf} {rgb}{0.20,0.20,0.20}
\definecolor{propcf}{rgb}{0.60,0.60,0.60}

\begin{figure}[tp]
\begin{center}%
\resizebox{\textwidth}{!}{%
\begin{timeline}[min=1930,max=2014,step=10,width=33.9cm,scale=0.8]
\tlbar[fadeout,color=propc]{1992}{1999}{PoliceSpeak}
\tlbar[fadeout,color=propc]{1982}{1991}{SEASPEAK}
\tlbar[id=icao,fadein,color=propc]{1981}{2014}{ICAO Phraseology}
\tlbar[id=caa,fadein,color=propc]{1981}{2014}{CAA Phraseology}
\tlbar[id=faa,fadein,color=propc]{1981}{2014}{FAA Air Traffic Control Phraseology}
\tlbar[color=propct]{2003}{2014}{\textit{Wycliffe Associates'} EasyEnglish}
\tlbar[fadein,color=propc]{1972}{2014}{Plain Language}
\tlbar[fadeout,color=propc]{1965}{2000}{E-Prime}
\tlbar[id=special,color=propc]{1959}{2014}{Special English}
\tlbar[id=basic,color=propc,labelabove]{1930}{2014}{Basic English}
\tlwconnect[color=propc]{basic}{special}
\tlbar[id=kisl,fadeout,color=propc]{1983}{1986}{Kodak International Service Language (KISL)}
\tlbar[id=case,fadein,fadeout,color=propc]{1984}{1988}{Clear And Simple English (CASE)}
\tlbar[id=cfe,color=propc]{1971}{1982}{Caterpillar Fundamental English (CFE)}
\tlconnect[color=propc]{basic}{cfe}
\tlwconnect[color=propc]{cfe}{kisl}
\tlconnect[color=propc]{cfe}{case}
\tlbar[id=cte,fadeout,color=propct]{1991}{2000}{Caterpillar Technical English (CTE)}
\tlconnect[color=propct]{cfe}{cte}
\tlbar[id=dce,fadeout,color=propct]{1996}{2003}{Diebold Controlled English (DCE)}
\tlconnect[color=propct]{cte}{dce}
\tlbar[id=kce,fadeout,color=propt]{1990.5}{1998}{Kant Controlled English (KCE)}
\tlconnect[color=propct]{kce}{cte}
\tlbar[id=clark,fadein,fadeout,color=propc]{1986}{1990}{\textit{Controlled English at Clark}}
\tlbar[id=rockwell,fadein,fadeout,color=propc]{1986}{1990}{\textit{Controlled English at Rockwell}}
\tlbar[id=hyster,fadeout,color=propc]{1987}{1990}{Hyster Easy Language Program (HELP)}
\tlbar[id=bull,color=propct]{1993}{1997}{Bull Global English}
\tlbar[id=nse,color=propc]{1996}{2000}{Nortel Standard English (NSE)}
\tlbar[id=smartp,fadein,color=propc]{1984}{2014}{SMART Plain English}
\tlconnect[color=propc]{cfe}{smartp}
\tlconnect[color=propc]{smartp}{clark}
\tlconnect[color=propc]{smartp}{rockwell}
\tlconnect[color=propc]{smartp}{hyster}
\tlwconnect[color=propct]{smartp}{bull}
\tlwconnect[color=propc]{smartp}{nse}
\tlbar[id=douglas,fadeout,color=propct]{1979}{1982}{\textit{Controlled English at Douglas}}
\tlwconnect[color=propct]{cfe}{douglas}
\tlbar[id=smartc,fadein,color=propct]{2005}{2014}{SMART Controlled English}
\tlwconnect[color=propct]{smartp}{smartc}
\tlbar[id=asd,color=propct]{2004}{2014}{ASD Simplified Technical English (ASD-STE)}
\tlconnect[color=propct]{asd}{smartc}
\tlbar[id=aecma,color=propc]{1986}{2004}{AECMA Simplified English (AECMA-SE)}
\tlconnect[color=propc]{douglas}{aecma}
\tlconnect[color=propct]{aecma}{asd}
\tlbar[id=boeing,fadeout,color=propc]{1998}{2001}{Boeing Technical English}
\tlconnect[color=propc]{aecma}{boeing}
\tlbar[id=ilsam,fadeout,color=propc]{1975}{1978}{International Language of Service and Maintenance (ILSAM)}
\tlconnect[color=propc]{cfe}{ilsam}
\tlconnect[color=propc]{ilsam}{aecma}
\tlbar[id=ibm,fadeout,color=propc]{1989}{1992}{\textit{Controlled English at IBM}}
\tlconnect[color=propc]{ilsam}{ibm}
\tlbar[id=easy,fadeout,color=propct]{1997}{2011}{\textit{IBM's} EasyEnglish}
\tlwconnect[color=propct]{ibm}{easy}
\tlbar[id=ee,fadeout,color=propc]{1983}{1986}{Ericsson English (EE)}
\tlconnect[color=propc]{ilsam}{ee}
\tlbar[id=mce,fadeout,color=propt]{1982}{1985}{Multinational Customized English (MCE)}
\tlconnect[color=propt]{ilsam}{mce}
\tlbar[id=pace,fadeout,color=propct]{1980}{1993}{Perkins Approved Clear English (PACE)}
\tlconnect[color=propct]{ilsam}{pace}
\tlbar[fadeout,color=propc]{1978}{1981}{NCR Fundamental English}
\tlbar[id=cogram,fadeout,color=propct]{1989}{1992}{COGRAM}
\tlbar[id=alcogram,fadeout,color=propct]{1992}{1995}{ALCOGRAM}
\tlconnect[color=propct]{cogram}{alcogram}
\tlbar[id=casl,color=propct]{1996}{2001}{Controlled Automotive Service Language (CASL)}
\tlbar[id=gm,fadeout,color=propct]{2000}{2003}{General Motors Global English}
\tlconnect[color=propct]{casl}{gm}
\tlbar[fadeout,color=propt]{2002}{2006}{Oc\'e Controlled English}
\tlbar[fadeout,color=propct]{2002}{2005}{Sun Proof}
\tlbar[fadeout,color=propt]{2002}{2010}{CLOUT}
\tlbar[fadeout,color=propct]{2003}{2007}{Avaya Controlled English}
\tlbar[color=propct]{2009}{2014}{Controlled Language for Crisis Management (CLCM)}
\tlbar[color=propct]{2009}{2014}{iHelp Controlled English (iCE)}
\tlbar[fadeout,color=propc]{2003}{2006}{\textit{Airbus Warning Language}}
\tlbar[color=propcf]{1990}{2014}{Standard Language (SLANG)}
\tlbar[color=propc]{2001}{2014}{\textit{Massachusetts Legislative Drafting Language}}
\tlbar[id=rulespeak,color=propcf]{1994}{2014}{RuleSpeak}
\tlbar[id=sbvr,color=propcf]{2005}{2014}{SBVR Structured English}
\tlwconnect[color=propcf]{rulespeak}{sbvr}
\tlbar[color=propf]{2013}{2014}{AIDA}
\tlbar[color=propf]{2010}{2014}{Voice Actions}
\tlbar[fadeout,color=propf]{1990}{2006}{ClearTalk}
\tlbar[fadeout,color=propf]{1999}{2003}{\textit{PROSPER Controlled English}}
\tlbar[fadeout,color=propf]{2005}{2010}{CPL}
\tlbar[id=clce,fadeout,color=propf]{2004}{2007}{Common Logic Controlled English (CLCE)}
\tlbar[id=ita,color=propf]{2010}{2014}{ITA Controlled English}
\tlconnect[color=propf]{clce}{ita}
\tlbar[color=propf]{2012}{2014}{Restricted English for Constructing Ontologies (RECON)}
\tlbar[fadeout,color=propf]{2004}{2011}{Restricted Natural Language Statement (RNLS)}
\tlbar[fadeout,color=propf]{2000}{2003}{\textit{MILE Query Language}}
\tlbar[id=clef,fadeout,color=propf]{2007}{2010}{\textit{CLEF Query Language}}
\tlbar[id=drafter,fadeout,color=propf]{1998}{2001}{\textit{Drafter Language}}
\tlconnect[color=propf]{drafter}{clef}
\tlbar[id=iconoclast,fadeout,color=propf]{1998}{2003}{\textit{ICONOCLAST Language}}
\tlbar[id=pills,fadeout,color=proptf]{2002}{2005}{\textit{PILLS Language}}
\tlconnect[color=proptf]{iconoclast}{pills}
\tlbar[color=propf]{2010}{2014}{\textit{Quelo Controlled English}}
\tlbar[color=propf]{1998}{2014}{Gellish English}
\tlbar[id=ginseng,color=propf]{2004}{2014}{\textit{Ginseng's Guided English}}
\tlbar[id=gino,fadeout,color=propf]{2006}{2009}{\textit{GINO's Guided English}}
\tlconnect[color=propf]{ginseng}{gino}
\tlbar[id=permis,color=propf]{2008}{2014}{PERMIS CNL}
\tlbar[id=clone,fadeout,color=propf]{2005}{2010}{CLOnE}
\tlconnect[color=propf]{clone}{permis}
\tlbar[id=atomate,fadeout,color=propf]{2010}{2013}{\textit{Atomate Language}}
\tlconnect[color=propf]{ginseng}{atomate}
\tlconnect[color=propf]{gino}{atomate}
\tlconnect[color=propf]{clone}{atomate}
\tlbar[id=ace,color=propf]{1996}{2014}{Attempto Controlled English (ACE)}
\tlconnect[color=propf]{ace}{atomate}
\tlbar[id=coral,color=propf]{2010}{2014}{Coral's Controlled English}
\tlconnect[color=propf]{ace}{coral}
\tlbar[id=bioquery,color=propf]{2009}{2014}{BioQuery-CNL}
\tlconnect[color=propf]{ace}{bioquery}
\tlbar[id=owlace,color=propf]{2006}{2014}{OWL ACE}
\tlconnect[color=propf]{ace}{owlace}
\tlbar[id=celt,fadeout,color=propf]{2003}{2011}{Controlled English to Logic Translation (CELT)}
\tlconnect[color=propf]{ace}{celt}
\tlbar[id=pengd,fadeout,color=propf]{2004}{2007}{PENG-D}
\tlbar[id=peng,color=propf]{2002}{2008}{PENG}
\tlconnect[color=propf]{ace}{peng}
\tlconnect[color=propf]{peng}{pengd}
\tlbar[id=penglight,color=propf]{2008}{2014}{PENG Light}
\tlconnect[color=propf]{peng}{penglight}
\tlbar[id=sos,fadeout,color=propf]{2007}{2011}{Sydney OWL Syntax (SOS)}
\tlconnect[color=propf]{peng}{sos}
\tlbar[fadeout,color=propf]{2008}{2011}{Rabbit}
\tlbar[fadeout,color=propf]{2005}{2010}{\textit{PathOnt CNL}}
\tlbar[id=e2v,fadeout,color=propf]{2001}{2004}{E2V}
\tlbar[id=lite,fadeout,color=propf]{2007}{2010}{Lite Natural Language}
\tlconnect[color=propf]{e2v}{lite}
\tlbar[id=dlengl,color=propf]{2009}{2014}{DL-English}
\tlconnect[color=propf]{lite}{dlengl}
\tlbar[color=propf]{2012}{2014}{SQUALL}
\tlbar[color=propf]{2012}{2014}{OWL Simplified English (OSE)}
\tlbar[color=propf]{2011}{2014}{ucsCNL}
\tlbar[color=propf]{2007}{2014}{Template Based Natural Language Specification (TBNLS)}
\tlbar[id=cpe,fadeout,color=propf]{1996}{2002}{Computer Processable English (CPE)}
\tlbar[id=clip,fadeout,color=propf]{2003}{2006}{Controlled Language for Inference Purposes (CLIP)}
\tlconnect[color=propf]{cpe}{clip}
\tlbar[fadein,fadeout,color=propf]{2000}{2004}{\textit{Sowa's Syllogisms}}
\tlbar[fadein,fadeout,color=propf]{2003}{2006}{First Order English}
\tlbar[fadeout,color=propf]{2003}{2007}{Pseudo Natural Language (PNL)}
\tlbar[fadeout,color=propf]{2000}{2010}{ForTheL}
\tlbar[color=propf]{2008}{2014}{Naproche}
\tlbar[color=propf]{2010}{2014}{Controlled Language of Mathematics (CLM)}
\tlbar[color=propf]{2012}{2014}{iLastic Controlled English}
\tlbar[color=propf]{2011}{2014}{\textit{OWLPath's Guided English}}
\tlbar[color=propf]{2011}{2014}{Gherkin}
\tlbar[fadeout,color=propf]{2002}{2005}{Formalized-English}
\draw[rounded corners=3mm,fill=white,thick] ($(0,\tlpos) + (0.3,2.5)$) rectangle +(16,13.5);
\node[anchor=west] at ($(0,\tlpos) + (0.8,14.7)$) {\sf\large Life spans:};
\fill[color=black!50!gray,draw,top color=white!20!gray,bottom color=black!20!gray,rounded corners=0.5mm,line width=0.15mm] ($(0,\tlpos) + (1,14)$) rectangle +(15mm,-6mm);
\fill[color=black!50!gray,draw,top color=white!20!gray,bottom color=black!20!gray,rounded corners=0.5mm,line width=0.15mm] ($(0,\tlpos) + (1.38,13)$) rectangle +(11.2mm,-6mm);
\fill[color=black!50!gray,draw,top color=white!20!gray,bottom color=black!20!gray,rounded corners=0.25mm,line width=0.15mm] ($(0,\tlpos) + (1.15,13)$) rectangle +(1.5mm,-6mm);
\fill[color=black!50!gray,draw,top color=white!20!gray,bottom color=black!20!gray,rounded corners=0.125mm,line width=0.15mm] ($(0,\tlpos) + (1,13)$) rectangle +(0.75mm,-6mm);
\fill[color=black!50!gray,draw,top color=white!20!gray,bottom color=black!20!gray,rounded corners=0.5mm,line width=0.15mm] ($(0,\tlpos) + (1,12)$) rectangle +(11.2mm,-6mm);
\fill[color=black!50!gray,draw,top color=white!20!gray,bottom color=black!20!gray,rounded corners=0.25mm,line width=0.15mm] ($(0,\tlpos) + (2.2,12)$) rectangle +(1.5mm,-6mm);
\fill[color=black!50!gray,draw,top color=white!20!gray,bottom color=black!20!gray,rounded corners=0.125mm,line width=0.15mm] ($(0,\tlpos) + (2.425,12)$) rectangle +(0.75mm,-6mm);
\node[anchor=west] at ($(0,\tlpos) + (2.8,13.7)$) {\sf\large period when language was studied or used};
\node[anchor=west] at ($(0,\tlpos) + (2.8,12.7)$) {\sf\large uncertain when language was first studied or used};
\node[anchor=west] at ($(0,\tlpos) + (2.8,11.7)$) {\sf\large uncertain whether or when language was discontinued};
\node[anchor=west] at ($(0,\tlpos) + (0.8,10.2)$) {\sf\large Influences:};
\draw[color=gray,ultra thick,line cap=round] ($(0,\tlpos) + (1.2,9)$) -- +(-0.15,0);
\draw[color=gray,ultra thick] ($(0,\tlpos) + (1.05,9)$) -- +(0,0.6);
\draw[color=gray,ultra thick,line cap=round] ($(0,\tlpos) + (1.2,8)$) -- +(-0.15,0);
\draw[color=gray,ultra thick,dotted] ($(0,\tlpos) + (1.05,8)$) -- +(0,0.6);
\node[anchor=west] at ($(0,\tlpos) + (1.5,9.2)$) {\sf\large reported influence};
\node[anchor=west] at ($(0,\tlpos) + (1.5,8.2)$) {\sf\large probable influence};
\node[anchor=west] at ($(0,\tlpos) + (0.8,6.7)$) {\sf\large Goals:};
\fill[color=black!50!propc,draw,top color=white!20!propc,bottom color=black!20!propc,rounded corners=0.5mm,line width=0.15mm] ($(0,\tlpos) + (1,6)$) rectangle +(15mm,-6mm);
\fill[color=black!50!propt,draw,top color=white!20!propt,bottom color=black!20!propt,rounded corners=0.5mm,line width=0.15mm] ($(0,\tlpos) + (1,5)$) rectangle +(15mm,-6mm);
\fill[color=black!50!propf,draw,top color=white!20!propf,bottom color=black!20!propf,rounded corners=0.5mm,line width=0.15mm] ($(0,\tlpos) + (1,4)$) rectangle +(15mm,-6mm);
\node[anchor=west] at ($(0,\tlpos) + (2.8,5.7)$) {\sf\large comprehensibility (\prop{c})};
\node[anchor=west] at ($(0,\tlpos) + (2.8,4.7)$) {\sf\large translation (\prop{t})};
\node[anchor=west] at ($(0,\tlpos) + (2.8,3.7)$) {\sf\large formal representation (\prop{f})};
\end{timeline}
}\end{center}
\caption{The timeline of the evolution of controlled English.}
\label{fig:timeline}
\end{figure}

The oldest CNL, Basic English, is also the most influential one. It influenced CFE, and indirectly ILSAM, both very influential languages in their own right. Altogether, more than 20 languages were directly or indirectly inspired by Basic English. Among the more recent languages, ACE is the most influential in terms of offspring languages.

Looking for an overall theme in the evolution of CNL, one can identify something that could be called three ``eras'': the general, technical, and logical eras. The \textbf{general era} lasted until the late 1960s or early 1970s. Only a few languages were defined and used during this time, all of which were designed to improve human comprehension and to serve as general languages with no specific application domain or narrow community in mind. These languages survived in their small niche, when during the subsequent \textbf{technical era} that began in the early 1970s, CNLs were applied to technical documentation for improved human comprehension as well as improved machine translation. Again, this branch of languages did not disappear at the end of the era and continues to be used today, but a new type of CNL emerged. During the \textbf{logical era} that began in the mid 1990s, many CNLs were created with some sort of mapping to formal logic, which enabled not only automatic processing but actual automatic interpretation. These three eras partly correspond to the three goals introduced in Section \ref{sec:types}: The first CNLs were of type \prop{c}, type \prop{t} emerged in the technical era, and type \prop{f} in the logical era.

\subsection{Evaluations}

Finally, we can turn to a crucial aspect that we have not yet discussed: Do CNLs actually achieve the goals they were designed for? A number of studies have been reported that evaluate the supposed advantages of these languages. The relevant research question obviously depends on the goal the language is supposed to achieve. In their most general forms, the research questions for the types \prop{c}, \prop{t}, and \prop{f} can be stated as follows:
\begin{itemize}
\item[\prop{c}] Does a CNL make communication among humans more precise and more effective?
\item[\prop{t}] Does a CNL reduce overall translation costs at a given level of quality?
\item[\prop{f}] Does a CNL make it easier for people to use and understand logic formalisms?
\end{itemize}
Each of these general research questions can be broken down, and most studies target more specific questions.

For type \prop{c}, two studies on AECMA-SE showed that the use of controlled English significantly improves text comprehension, with a particularly large effect for complex texts and non-native speakers \cite{chervak1996faaaam,shubert1995twc}. The results of other studies were similar but not significant \cite{steward1998master}. The language CLCM has been found to have a positive effect on reading comprehension for most groups of readers under certain circumstances such as stress situations \cite{temnikova2012phd}.

Concerning type \prop{t}, it has been reported that the use of the controlled language MCE for machine-assisted translation leads to a ``five-to-one gain in translation time'' \cite{ruffino1982pemt}.
Similar results have been presented for the language PACE, with which post-editing of machine-assisted translation is ``three or four times faster'' than without \cite{pym1990tatc}.
It has been shown that the adherence to typical CNL rules improves post editing productivity and machine translation quality \cite{aikawa2007mt,obrien2007mt}.
For the language CLCM, it has been reported that CNL texts are easier to translate than uncontrolled ones \cite{temnikova2009ismtcl,temnikova2012phd} and that the time needed for post-editing is reduced on average by 20\% \cite{temnikova2010lrec,temnikova2012phd}.

Studies on type \prop{f} can be subdivided into those that test the general usability of CNL tools and those that specifically evaluate the comprehensibility of the actual languages. Starting with the usability studies, it has been shown for the language CLOnE that its interface is more usable than a common ontology editor \cite{funk2007iswc}. Similarly, Coral's controlled English has been shown to be easier to use than a comparable common query interface \cite{kuhn2012corpora}. Positive usability results for CNL tools have also been reported for GINO \cite{bernstein2006iswc}, CLEF \cite{hallett2007coli}, CPL \cite{clark2007kcap}, PERMIS \cite{inglesant2008soups}, Rabbit \cite{dimitrova2008iswc}, and ACE \cite{kuhn2009semwiki}. Turning to the comprehensibility studies, it has been shown for the CLEF query language that common users are able to correctly interpret given statements \cite{hallett2007coli}. ACE has been shown to be easier and faster to understand than a common ontology notation \cite{kuhn2013swj}, whereas experiments on the Rabbit language gave mixed results \cite{hart2008eswc}.

In addition to these high-level evaluations, more specific tests have been reported such as evaluations on coverage \cite{bernstein2006jena,kaljurand2007phd}, performance, convergence \cite{adriaens1995tmi}, parseability \cite{wojcik1993acl}, computational complexity \cite{pratthartmann2003jolli,thorne2009cnlmain}, text complexity, and text length \cite{temnikova2012phd}.

In general, there seems to be good evidence for each of the language types that the use of CNL can be advantageous. This does not mean, of course, that CNL approaches always perform better. This depends heavily on the precise problem domain, the background of the users, and --- perhaps most importantly --- the quality of the design of the language and its supporting tools.

\section{Conclusions}
\label{sec:conclusions}

To conclude, we can come back to the aims set out in the introduction of this article. The first goal was to get a better theoretical understanding of the nature of controlled languages. First of all, this article shows that despite the wide variety of existing CNLs, they can be covered by a single definition. The criteria of the proposed definition include virtually all languages that have been called CNLs in the literature. We could show that these languages form a widely scattered but connected cloud in the conceptual space between natural languages on the one end and formal languages on the other. The informal statement that CNLs are more formal than natural languages but more natural than formal ones is substantiated and verified.

The next goal was to establish a common terminology and a common model. We emphasized the difference between characteristics of the \emph{environments} of languages on the one hand and the properties of the \emph{languages themselves} on the other. Both aspects are important, but the second is more difficult to capture in a quantitative way. Nine general properties have been collected to describe the application environments of CNLs. As a novel addition to this model, we proposed the four-dimensional PENS scheme to describe inherent language properties. This scheme allows for classification of CNLs on a discrete scale on the dimensions of precision, expressiveness, naturalness, and simplicity. Together, this allows us to formally model the important properties of languages and their environments in a simple way, and to put order and structure to a previously fuzzy and disconnected field.

The third goal was to provide a starting point for researchers interested in CNL. The most important conclusion in this respect is the fact that many more CNLs exist than have been found in any previous survey. Previously, the most comprehensive overview counted 41 CNLs \cite{pool2006claw} based on various natural languages, whereas this survey covers 100 languages for English alone. The diversity of languages and the different environments in which they were studied and used apparently had the consequence that many CNL researchers and developers were not aware of a large number of relevant languages. As a starting point for researchers, this work presents a diverse sample of twelve important and influential languages, along with a long list of all CNLs collected. The introduced model of languages and environments can also facilitate the identification of a particular research focus and the collection of relevant prior work.

The fourth goal was to help CNL developers make design decisions. To that aim, the data of this survey can be used to direct developers to existing CNL approaches in a given environment and problem domain. The data can reveal whether a certain kind of CNL usage is common, rare, or inexistent until now, which can be used as an indication of the amount of original work required. Furthermore, the typical language properties of CNLs in terms of precision, expressiveness, naturalness, and simplicity can be retrieved for a given usage scenario. This information might be very useful to identify important design decisions and to find existing approaches to build upon.

I would like to conclude with the observation that the study of controlled languages is a very dynamic and highly interdisciplinary field, for the most part occupying small niches in the academic, industrial, and governmental worlds. However, adding all these niches together gives us a large body of past and ongoing work. Assuming that people will have to interact even more closely with computers and across language borders in the future, I am convinced that we will see even much more work in this area.

\appendix

\appendixsection{Full List of English-based Controlled Natural Languages}

Below is the full list of {\numberofcnls} English-based CNLs in alphabetical order. See Section \ref{sec:languages} for the details of this collection.

\langparax{AECMA Simplified English (AECMA-SE)}{\pens2551, \prop{c~w~d~i}}{ \cite{aecma1986se}
was the predecessor of ASD Simplified Technical English. \emph{See Section \ref{sec:cnls}}.}

\langparax{AIDA}{\pens2541, \prop{f~w~a}}{ \cite{kuhn2013eswc} is a CNL to allow for informal and underspecified representations of scientific assertions in an approach for semantic publishing called ``nano\-pub\-li\-ca\-tions.'' Single English sentences are used as a scaffold for underspecified representations and for the inclusion of informal statements in formal RDF-based structures. These sentences are Atomic, Independent, Declarative, and Absolute (hence the name AIDA). This is an example:
\langexample{The degree of hepatic reticuloendothelial function impairment does not differ between cirrhotic patients with and without previous history of SBP.}
}

\langparax{``Airbus Warning Language''}{\pens2133, \prop{c~w~d~i}}{ \cite{spaggiari2003eamtclaw} is a language for short industrial warnings, focusing on abbreviations and restricting the word order. This is an exemplary statement:
\langexample{ENG1 REV NOT LOCKED}
}

\langparax{ALCOGRAM}{\pens2551, \prop{c~t~w~d~a~i}}{ \cite{adriaens1992coling} is a CNL developed at Alcatel. It originated from COGRAM as an ``algorithmic variant,'' focusing on the use within a computer-aided language learning tool. In contrast to COGRAM, which consists of three components that declaratively define the language, ALCOGRAM is defined based on a four-staged algorithm. Each of these four stages checks certain aspects: preparatory textual control (e.g., ``Define technical terms and acronyms in advance''), syntactic control (e.g., ``Write one instruction per sentence for single actions''), lexical control (e.g., ``Avoid gender-specific language''), and micro control (e.g., ``Use words for a number when it is the first word in the sentence''). These are two examples of ALCOGRAM sentences:
\langexample{Set the switch to the middle. Press the button on your right.}
\langexample{When the test circuit is called, a test tone with the proper transmit level is returned.}
}

\langparax{ASD Simplified Technical English (ASD-STE).}{\pens2551, \prop{c~t~w~d~i}}{
\emph{See Section \ref{sec:cnls}.}
}

\langparax{``Atomate Language''}{\pens4243, \prop{f~w~d~a}}{ \cite{vankleek2010www} is part of the Atomate interface, which lets users define simple automatic tasks and reminders taking context and current activity into account. The language was inspired by CLOnE, ACE, and the GINO and Ginseng systems. This is an example of such a task definition:
\langexample{Alert me when my location is home on/after Tuesdays at 5pm with the message: Trash day!}
A special editor supports users in writing such sentences, using a mixture of predictive editing and conceptual authoring. The sentences are mapped to RDF and automatically triggered when the preconditions are met.
}

\langparax{Attempto Controlled English (ACE).}{\pens4343, \prop{f~w~a}}{
\emph{See Section \ref{sec:cnls}.}
}

\langparax{Avaya Controlled English}{\pens2551, \prop{c~t~w~d~i}}{ \cite{avaya2004styleguide} is a language for technical publications in the telecommunication and computing industry. Its use should reduce translation costs and should make texts easier to understand for human readers. It puts restrictions on the lexicon (e.g., ``Do not use \emph{abort}''), grammar (e.g., ``Use active voice''), semantics (e.g., ``Use \emph{may} only to grant permission''), and style (e.g., ``Put command names in bold monospaced type''). An open list of about 250 words defines preferred terminology for the given computer and telephony domain, and clarifies usage and meaning of these words. These are two examples of sentences:
\langexample{This procedure describes how to connect a dual ACD link to the server.}
\langexample{If the primary server fails, you can use the secondary server.}
}

\langparax{Basic English.}{\pens2551, \prop{c~w}}{
\emph{See Section \ref{sec:cnls}.}
}

\langparax{BioQuery-CNL}{\pens5244\todo{?}, \prop{f~w~d~a}}{ \cite{erdem2009bionlp} is a language for biomedical queries. It serves as an interface language for a query engine based on answer set programming. BioQuery-CNL was initially designed as a subset of ACE with some small modifications handled in a preprocessing step. The ACE parser was used for processing the language. In later versions, however, the language diverged from ACE and evolved into an independent language with its own parser. This is an exemplary query:
\langexample{What are the genes that are targeted by all the drugs that belong to the category Hmg-coa reductase?}
}

\langparax{Boeing Technical English}{\pens2551, \prop{c~w~i}}{ \cite{wojcik1998claw} was an extension of AECMA Simplified English to improve readability and consistency of documents, with the specific goal to broaden the scope beyond the aviation domain. The language seems to have been discontinued and apparently never came to be deployed at Boeing.}

\langparax{Bull Global English}{\pens2551, \prop{c~t~w~d~i}}{ \cite{smart1994mtnews} or \textbf{Bull Controlled English} is a language developed at Groupe Bull, a French computer company.
It was probably influenced by SMART Plain English. Bull Global English can be summarized by the following ten rules \cite{karkaletsis1997clsr}, which have a considerable overlap with the rules of Caterpillar Fundamental English:
\begin{center}\small
\begin{minipage}[t]{0.4\textwidth}
1. Make positive statements.\\
2. Keep sentence length 21 words.\\
3. Avoid false nomenclature.\\
4. One thought per sentence.\\
5. Use simple sentence structures.
\end{minipage}
\begin{minipage}[t]{0.47\textwidth}
6. Use active voice and parallel construction.\\
7. Avoid conditional tenses.\\
8. Avoid abbreviations and colloquialisms.\\
9. Use correct punctuation.\\
10. Use standardised nomenclature.
\end{minipage}
\end{center}
}

\langparax{CAA Phraseology}{\pens2132, \prop{c~s~d~g}}{ \cite{caa2011rt} is a language for air traffic control introduced by the Civil Aviation Authority (CAA) in the 1980s or possibly earlier. It is very similar to the phraseologies by FAA and ICAO.}

\langparax{Caterpillar Fundamental English (CFE).}{\pens2551, \prop{c~w~d~i}}{
\emph{See Section \ref{sec:cnls}.}
}

\langparax{Caterpillar Technical English (CTE)}{\pens2551, \prop{c~t~w~d~i}}{ \cite{hayes1996claw,kamprath1998claw} is the second CNL developed at Caterpillar. Its development started in 1991, that is, almost a decade after the discontinuation of CFE. Apart from improving consistency and reducing ambiguity of technical documentation, the goal of CTE was to improve translation quality and reduce translation costs with the help of machine translation. This is an example of a CTE text:
\langexample{This category indicates that an alternator is malfunctioning. If the indicator comes on, drive the machine to a convenient stopping place. Investigate the cause and determine the solution.}
In contrast to CFE, texts in CTE are supposed to be translated before given to personnel in non-English speaking countries. As a further difference, CTE was designed to be an ``enforceable controlled English'' that comes with an authoring tool that enforces the compliance with the restrictions. The CTE lexicon consists of about 70,000 terms with a ``narrow semantic scope'' (compared to CFE's less than 1,000 terms with a broader semantic scope). The syntax is restricted too, including restrictions on the use of conjunctions, pronouns, and subordinate clauses. CTE comes with a language checker that allows for interactive disambiguation on the lexical level, enriches the technical texts with SGML annotations, and uses the syntax analyzer of the KANT system (see KANT Controlled English below).}

\langparax{Clear And Simple English (CASE)}{\pens2551, \prop{c~w~d~i}}{ \cite{pym1990tatc} was a controlled English introduced in the 1980s at the J.I. Case Company, a manufacturer of construction and agricultural equipment. It descended from CFE.}

\langparax{ClearTalk}{\pens3333, \prop{f~w~a}}{ \cite{skuce2003cleartalk} is a CNL for the Semantic Web first presented in the 1990s. Its creator claims that documents in ClearTalk can be ``almost automatically'' translated into a formal logic notation and into other natural languages. It ``offers a flexible degree of formality'' that lets an author choose to ``leave or remove ambiguity.'' It has been used to encode more than 25,000 facts in different technical domains. ClearTalk is heavily restricted on the syntactic level (e.g., basic sentences have the general form \emph{subject predicate complement modifier-phrases}) as well as on the semantic one (e.g., the determiner \emph{a} at subject position represents universal quantification). These restrictions are expressed in a large number of rules.
Two examples of sentences are shown here:
\langexample{Any adverb that modifies a verb must be adjacent to (that verb or another adverb).}
\langexample{Mary hopes that [- Bill loves her -].}
ClearTalk can itself be described in ClearTalk; the first example is from this self-description. Different forms of parentheses are used to disambiguate different kinds of scopes.
}

\langparax{``CLEF Query Language''}{\pens5143, \prop{f~w~d~a}}{ \cite{hallett2007coli} is a language used within a system called CLEF (Clinical E-Science Framework), which should help clinicians, medical researchers, and hospital administrators to query electronic health records. The language was influenced by the Drafter language. Basic queries are composed of three elements: the set of relevant patients, the received treatments, and the outcomes. This is an example:
\langexample{For all patients with cancer of the pancreas, what is the percentage alive at five years for those who had a course of gemcitabine?}
Complex queries can have multiple elements of the same type. The system employs a conceptual authoring approach for writing queries, which are then translated in several steps to SQL and given to a database engine.
}

\langparax{COGRAM}{\pens2551, \prop{c~t~w~d~a~i}}{ \cite{adriaens1992coling} was a controlled language developed in the late 1980s for the telecommunication domain (at Alcatel). It was developed as a response to the finding that the existing controlled languages AECMA Simplified English, Ericsson English, and IBM's controlled English were ``incomplete and defective in many ways.'' COGRAM consists of a vocabulary of approximately 5,000 words plus another 1,000 technical terms, and a grammar with about 150 rules. These rules fall into three categories: ``Do not use X,'' ``Use only X,'' and ``Avoid (try not to use) X.'' Grammar rules of the last type can be seen as style-guides that do not restrict the coverage of the language. The language definition is divided into three components: lexical (e.g., ``Use short infinitives of regular action verbs''), syntactic (e.g., ``Do not use a participle to introduce an adverbial clause''), and stylistic (e.g., ``Expound major topics, restrict minor topics'').
The definition of COGRAM was found to be ``not the most motivating of texts for technical writers to use in the writing process,'' which led to the development of ALCOGRAM.}

\langparax{Common Logic Controlled English (CLCE)}{\pens5333, \prop{f~w~a}}{ \cite{sowa2004clce} is a language that can be translated into first-order logic with equality in the form of the Conceptual Graph Interchange Format (CGIF). It is defined by a grammar in Backus-Naur form ``that allows every ambiguity to be resolved when a sentence is parsed.''
Some of the most important syntax restrictions are: no plural nouns, only present tense, and variables instead of pronouns. For an unambiguous mapping to logic, a number of interpretation rules are applied
and parentheses are used to determine the structure of deeply nested sentences.
Sentences in this language should be similar to those found in software documentation and textbooks of mathematics, for example:
\langexample{If some person x is the mother of a person y, then the person y is a child of the person x.}
\langexample{
Declare give as verb (agent gives recipient theme)
(agent gives theme to recipient)
(theme is given recipient by agent)
(theme is given to recipient by agent)
(recipient is given theme by agent).
}
Imperative sentences, as the second example, are used to import or declare words. Names, nouns, verbs, adjectives, adverbs, and prepositions can be declared in this way.
}

\langparax{Computer Processable English (CPE)}{\pens5243, \prop{f~w~a}}{ \cite{pulman1996claw,sukkarieh1999iwcs} is a controlled language that can be ``completely syntactically and semantically analysed.'' An early version of the language used KIF (Knowledge Interchange Format) as its logic formalism, whereas McLogic was used later on.
The language comes with a bidirectional grammar implemented as a Prolog unification grammar.
Two examples are shown here:
\langexample{Every animal X eats some animal that is smaller than X.}
\langexample{Every registered user who has borrowed less than ten copies can borrow every available copy.}
The mapping to logic seems to be deterministic, even though the available literature is not explicit about this.
}

\langparax{Computer Processable Language (CPL)}{\pens3342, \prop{f~w~i}}{ \cite{clark2005flairs} is a controlled variant of English developed at Boeing. It is very different from earlier CNL approaches where Boeing was involved in, such as ASD-STE and Boeing Technical English. CPL is much more restricted than these earlier approaches and sacrifices to some degree expressiveness and naturalness for the sake of automated reasoning support. Basic CPL sentences are restricted to the pattern \emph{subject + verb + complements + adjuncts}. There are further restrictions on the syntax, for example that definite references have to be used instead of pronouns.
Statements involving universal quantification are constructed from seven templates such as ``If sentence1 then typically sentence2,'' where \emph{sentence1} and \emph{sentence2} are basic CPL sentences of the structure introduced above and where \emph{typically} is a reliability degree: one of \emph{(almost) always}, \emph{usually}, \emph{sometimes}, and \emph{never}. These are two examples of CPL sentences:
\langexample{IF a person is carrying an entity that is inside a room THEN (almost) always the person is in the room.}
\langexample{AFTER a person closes a barrier, (almost) always the barrier is shut.}
A parser translates CPL sentences into a frame-based language with well-defined semantics. In contrast to most other logic-based CNL approaches with custom-built parsers, the parsing process of CPL involves different external tools and resources. An existing parser for unrestricted English is used to generate an intermediary logical form. Then, WordNet and other resources are used to make a ``best guess.'' The resulting logical representation is then paraphrased and shown to the user for verification or correction.
}

\langparax{Controlled Automotive Service Language (CASL)}{\pens2551, \prop{c~t~w~d~i}}{ \cite{means1996claw,means2000claw} is a controlled language for writing service manuals and bulletins at General Motors developed in the 1990s.
The goal was to improve translatability, as well as consistency and readability. The approach moved from an ``author-centric model'' towards a ``hybrid model'' that included the role of an editor, before it went to full production in 2000 \cite{godden2000claw}. The CASL restrictions are defined by 62 rules, including restrictions on sentence structure, word order, vocabulary, and punctuation. This is an exemplary sentence:
\langexample{Several diseases result from asbestos exposure, with latency periods of 10 to 40 years or longer.}
Writers are supported by a software tool called CASLChecker.}

\langparax{``Controlled English at Clark''}{\pens2551, \prop{c~w~d~i}}{ \cite{adriaens1992coling} was a language used at the Clark Material Handling Company.
It was developed around the late 1980s and was influenced by SMART Plain English.}

\langparax{``Controlled English at Douglas''}{\pens2551, \prop{c~t~w~d~i}}{ \cite{kleinman1982tc} was a language developed in 1979 by the McDonnell Douglas aerospace company for their technical manuals. It was based on a dictionary of about 2,000 words (most of them verbs), favoring short and simple words and aiming at a single word per meaning and a single meaning per word. In addition to the words of the dictionary, ``nomenclature words'' can be introduced.
The goal was to improve readability, translatability, and standardization. It was probably influenced by CFE and had itself an influence on AECMA SE.}

\langparax{``Controlled English at IBM''}{\pens2551, \prop{c~\todo{t?~}w~d~i}}{ \cite{adriaens1992coling} was a language developed and used at IBM in the late 1980s. It was influenced by ILSAM and might have influenced EasyEnglish, which was also developed at IBM several years later. It relied on a closed list of words, and writers were assisted by different instruction programs.
}

\langparax{``Controlled English at Rockwell''}{\pens2551, \prop{c~w~d~i}}{ \cite{adriaens1992coling} was a language used at the company Rockwell International. It was developed around the late 1980s and was influenced by SMART Plain English.}

\langparax{Controlled English to Logic Translation (CELT)}{\pens4243, \prop{f~w~i}}{ \cite{pease2010ontology} is a controlled natural language presented in 2003. It is a domain-independent language inspired by ACE. In contrast to ACE, it uses existing linguistic and ontological resources, concretely the SUMO ontology and WordNet. These are two exemplary sentences:
\langexample{Dickens writes Oliver Twist in 1837.}
\langexample{Every boy likes fudge.}
The syntax structure of CELT sentences is deterministically parsed. Heuristics are applied only afterwards to map the words to SUMO and WordNet.
The language is implemented as a unification grammar in Prolog.
}

\langparax{Controlled Language for Crisis Management (CLCM)}{\pens2551, \prop{c~t~w~d~a}}{ \cite{temnikova2010lrec,temnikova2011ranlp,temnikova2012phd} is a language for writing instructions about how to deal with crisis situations. The language is defined by about 80 simplification rules. These simplification rules include restrictions on text structure (e.g., ``Write a title for every specific situation''), formatting (e.g., ``Separate with a new line each block of instructions''), lexicon (e.g., ``avoid technical terms''), syntax (e.g., ``Avoid passive voice''), semantics (e.g., ``Use only literal meaning''), and pragmatics (e.g., ``Remove unimportant information'').
}

\langparax{Controlled Language for Inference Purposes (CLIP)}{\pens5243, \prop{f~w~a}}{ \cite{sukkarieh2003eamtclaw} is a language based on the logic notation McLogic and influenced by CPE. It is ``semantically-driven,'' meaning that it was designed around the given logic formalism and not vice versa. Two examples are shown here:
\langexample{Every student who laughs succeeds}
\langexample{Smith and Jones sign five contracts}
}

\langparax{Controlled Language for Ontology Editing (CLOnE)}{\pens5244, \prop{f~w~a}}{ \cite{funk2007iswc}, previously called \textbf{CLIE Controlled Language}, is a CNL designed as a front-end language for OWL, covering only a small subset of it. It is defined by ten basic sentence patterns. It adds procedural semantics on top of OWL to introduce and remove entities and axioms. These are two examples of CLOnE sentences:
\langexample{Persons are authors of documents.}
\langexample{Forget everything.}
}

\langparax{Controlled Language Optimized for Uniform Translation (CLOUT)}{\pens2551, \prop{t~w~i}}{ \cite{muegge2007clientside} is a CNL to improve machine translation. It puts restrictions on the vocabulary and prohibits structures such as passive voice and pronouns.
}

\langparax{Controlled Language of Mathematics (CLM)}{\pens5333, \prop{f~w~d~a}}{ \cite{humayoun2010rcs} is a language for expressing mathematical texts, as found in textbooks.
The language is similar to Naproche CNL and ForTheL. The grammar of CLM is implemented in GF (Grammatical Framework) and allows for deterministic translation into first-order logic. The goal is to automatically verify mathematical proofs.}

\langparax{Coral's Controlled English}{\pens5144, \prop{f~w~d~a}}{ \cite{kuhn2012corpora} is a controlled language for expressing formal queries to annotated text corpora. It is influenced by ACE, but is much less expressive, simpler and more domain-specific. It is embedded into a query interface called Coral to enable users with no particular background in computer science to effectively use large corpora of annotated texts. This is an exemplary query:
\langexample{Find all passages where a noun phrase contains a verb phrase; the verb phrase precedes a prepositional phrase; the prepositional phrase contains a verb ``see'';}
Such queries are deterministically mapped to AQL, an existing formal query language.
The language is defined by 51 simple grammar rules.}

\langparax{Diebold Controlled English (DCE)}{\pens2551, \prop{c~t~w~d~i}}{ \cite{hayes1996claw,moore2000claw} is a controlled language developed at Diebold with the goal to make translation faster and less expensive by assisting human translators with specific translation tools. It was inspired by CTE, but is less strict concerning lexicon and grammar, making the approach more flexible. It consists of three main components: a lexical database, a set of grammar rules, and a checking tool.}

\langparax{DL-English}{\pens5244, \prop{f~w~a}}{ \cite{thorne2009cnlmain} is a Description Logic based controlled language presented together with other similar languages to study and compare their computational complexity. It is similar to Lite Natural Language by the same research group.}

\langparax{``Drafter Language.''}{\pens4143, \prop{f~w~d~a}}{
\emph{See Section \ref{sec:cnls}.}
}

\langparax{E-Prime}{\pens1551, \prop{c~w~a}}{ or \textbf{E'.}
\emph{See Section \ref{sec:cnls}.}
}

\langparax{E2V.}{\pens5244, \prop{f~w~a}}{
\emph{See Section \ref{sec:cnls}.}
}

\langparax{EasyEnglish (by IBM)}{\pens1551, \prop{c~t~w~i}}{ \cite{bernth1997anlp}, not to be confused with Wycliffe Associates' EasyEnglish, is a language developed at IBM, which might have been influenced by an earlier controlled English at the same company \cite{adriaens1992coling}. The main goal of EasyEnglish was to improve machine translation. The approach is based on a sophisticated grammar checker that returns suggestions and warnings. Apart from detecting common grammar errors, the system can enforce the use of a certain controlled vocabulary and can spot ambiguities. For such ambiguities, the system can propose alternatives, but it is ultimately up to the user whether to follow the system's suggestions or not. The problems encountered in a given document are quantified in the form of a clarity index, which must be above a certain threshold value. The fact that the restrictions of the language are not enforced but just suggested does not make the language more precise or simpler than full natural English. EasyEnglish has been extended later to check not only on the sentence level but also on the document level, and this has been implemented in a tool called EasyEnglishAnalyzer \cite{bernth2006claw}.}

\langparax{EasyEnglish (by Wycliffe Associates)}{\pens2551, \prop{c~t~w~d}}{ \cite{betts2003eamtclaw}, not to be confused with IBM's EasyEnglish, is a controlled language used for transcribing biblical texts. The original goal was to improve the translation process into other languages, but EasyEnglish is also directly used by readers with limited knowledge of English. The language is restricted with respect to lexicon, syntax, and semantics. There are two levels: Level A makes use of about 1,200 words, whereas level B has a larger lexicon of about 2,800 words. In either case, the meaning of these words is restricted. For example, \emph{fair} can only mean \emph{unbiased}, and \emph{to see} cannot be used in the sense \emph{to meet}. It is possible to use words that are not on the list, if they are explained in separate EasyEnglish sentences. The following is an excerpt of a text in EasyEnglish (\emph{moor} is not in the lexicon and has to be explained):
\langexample{
The Highlands of Scotland consist of lakes, mountains and moors. The moors are flat empty lands where no trees grow. This land is wonderful and magnificent because it is so empty.
}
There is a strict sentence length limit of 20 words, and paragraphs may not contain more than 150 words. Sentence structure is kept simple by allowing not more than two finite clauses and not more than two prepositional phrases per sentence. Furthermore, deep nesting and passives are restricted. In addition, texts should adhere to logical simplicity: ``EasyEnglish writers are encouraged to identify the basic idea units in a complex sentence or paragraph and arrange them in logical order.''
}

\langparax{Ericsson English (EE)}{\pens2551, \prop{c~\todo{t?~}w~d~i}}{ \cite{adriaens1992coling} was a language developed at Ericsson in the early 1980s, influenced by ILSAM. It is built on a closed list of acceptable words, but other words can be introduced if accompanied by a definition in EE.
}

\langparax{FAA Air Traffic Control Phraseology.}{\pens2132, \prop{c~s~d~g}}{
\emph{See Section \ref{sec:cnls}.}
}

\langparax{First Order English}{\pens5343, \prop{f~w~a}}{ \cite{pool2006claw} is a controlled natural language that maps to first-order logic.
No detailed description of this language is available.
}

\langparax{Formalized-English (FE).}{\pens5433, \prop{f~w~a}}{
\emph{See Section \ref{sec:cnls}.}
}

\langparax{ForTheL}{\pens5333, \prop{f~w~d~a}}{ \cite{vershinin2000ijita} is a CNL for mathematical texts similar to Naproche CNL and CLM. The name stands for ``Formal Theory Language.'' Statements in this language can be automatically translated into first-order logic with equality. The following is an exemplary text:
\langexample{
Lemma 1. Each set has a subset.\\
Proof. 0 is a subset of all sets. QED.
}}

\langparax{Gellish English}{\pens4243, \prop{f~w~a~i}}{ \cite{vanrenssen2005phd} is a controlled language designed as a common data language for industry. The first version was ready in 1998. Basically, it consists of simple subject--predicate--object structures with predefined relations in the form of fixed phrases such as ``is a specialization of'' and ``is valid in the context of.'' These are two examples:
\langexample{collection C each of which elements is a specialization of animal}
\langexample{
the Eiffel tower has aspect h1\\
h1 is classified as a height\\
h1 is qualified as 300 m
}
Meta-information about the context of such statements can be expressed in the form of additional ``accessory facts.'' Gellish builds upon a fixed upper ontology with a large number of predefined concepts and relation types. Texts in Gellish can be transformed into a formal tabular representation.
The semantics of the language is not fully formalized, which means that there is no mapping to an established logic formalism. Gellish support simple kinds of if--then rules \cite{vanrenssen2011fomi}, but these rules do not allow for universal quantification over several variables in a general way.
}

\langparax{General Motors Global English}{\pens2551, \prop{c~t~w~d\todo{?}~i}}{ \cite{means2000claw} or just \textbf{Global English} is a controlled language developed at General Motors. The goal was to improve comprehension for non-native speakers and translatability. It is defined by 15 rules based on four principles: ``be brief,'' ``be clear,'' ``be direct,'' and ``be culturally alert.'' These rules include a limit on the sentence length and grammatical restrictions such as the exclusion of passive voice. The language evolved from a reduced set of twelve of the 62 rules of the CASL language, which was developed at General Motors several years earlier. In contrast to CASL, Global English does not come with a software tool for checking the compliance with the restrictions.}

\langparax{Gherkin}{\pens5343, \prop{f~w~d~a}}{ \cite{necas2011master} is a language for writing executable scenarios for software specifications. This is an excerpt of a scenario description:
\langexample{
Scenario: Unsuccessful registration due to full course\\
\phantom{-- }Given I am a student\\
\phantom{-- }And a lecture ``PA042'' with limited capacity of 20 students\\
\phantom{-- }But the capacity of this course is full\\
\phantom{-- }[...]
}
The structuring words such as \emph{Given}, \emph{And}, and \emph{But} are fixed. The restrictions on the remaining text such as ``I am a student'' are implemented in ordinary programming languages using regular expressions, and are stored in small modules called ``step definitions.'' The concrete step definitions are not part of Gherkin, but have to be implemented for the particular task at hand. Gherkin is therefore highly customizable and extensible, and the classification given here is meant to apply to a typical concrete language that is based on Gherkin.
}

\langparax{``GINO's Guided English''}{\pens4243, \prop{f~w~a}}{ \cite{bernstein2006iswc} is a language used in GINO, a system to query and edit ontologies. The language was influenced by Ginseng and supports the same kinds of queries. In addition, GINO has some limited support for procedural statements to introduce new entities, for instance:
\langexample{There is a subclass of class water area named lake.}
Query statements are mapped to SPARQL and procedural statements map to OWL axioms to be added or modified. Queries can exhibit structural ambiguity, in which case the system evaluates all possible interpretations and shows to the user the union of their answers. The grammar that describes the language consists of 120 grammar rules.}

\langparax{``Ginseng's Guided English''}{\pens5143, \prop{f~w~a}}{ \cite{bernstein2006jena}
is a CNL used in a system called Ginseng, which is a query interface to access knowledge bases in the form of OWL ontologies. The vocabulary for the language is loaded from the respective ontologies. These are two examples of queries:
\langexample{What are the capitals of states that border Nevada?}
\langexample{Is there a city that is the highest point of a state?}
The grammar consists of 120 static grammar rules plus additional dynamic rules generated from the ontologies. \todo{ambiguity?}}

\langparax{Hyster Easy Language Program (HELP)}{\pens2551, \prop{c~w~d~i}}{ \cite{smart2003wft} is a controlled English developed in the 1980s for maintenance manuals for lift trucks. It is based on SMART Plain English and thus indirectly on CFE \cite{pym1990tatc}.}

\langparax{ICAO Phraseology}{\pens2132, \prop{c~s~d~g}}{ \cite{eurocontrol2009phraseology} is controlled language for air traffic control defined by the International Civil Aviation Organisation (ICAO) in the 1980s or even earlier. It is very similar to the phraseologies by FAA and CAA.}

\langparax{``ICONOCLAST Language''}{\pens4353, \prop{f~w~d~a}}{ \cite{power1999ewnlg} is a CNL to write patient information leaflets. It is similar to the Drafter language. A conceptual authoring approach is employed and a formal logic representation is used in the background. This is a simple example:
\langexample{If you develop a rash, you should consult your doctor.}
}

\langparax{iHelp Controlled English (iCE)}{\pens2551, \prop{c~t~w~i}}{\footnote{\url{http://www.lindy-hop.co.uk/iHelp/ice/}} is a language developed by iHelp Ltd, a documentation consultancy company. iCE consists of ``a set of flexible rules and vocabularies for companies wishing to standardise and improve their information.''}

\langparax{iLastic Controlled English}{\pens5343, \prop{f~w~i}}{ \cite{ilastic2012doc} is a language to allow non-developers to write intuitive and natural scripts that automatically retrieve, transform, and combine data from the web, databases, files, and other resources. This is an exemplary statement:
\langexample{delete all files under the tmp folder if the space of the disk is lower than 1024.}
}

\langparax{International Language of Service and Maintenance (ILSAM)}{\pens2551, \prop{c~w~d~i}}{ \cite{pym1990tatc}
is an influential language similar to Caterpillar Fundamental English, from which it was derived in the 1970s.
}

\langparax{ITA Controlled English (ITA CE)}{\pens3333, \prop{f~w~\todo{d?~}i}}{ \cite{mott2010ita} is a controlled language defined by the International Technology Alliance, a US/UK military research program. It is inspired by CLCE, but is less strict in terms of precision: It has an ``informal meaning and a semi-formal mapping to predicate logic.'' The following are two examples of statements of different types:
\langexample{if ( the person X has the person Y as brother ) and ( the person Z has the person X as father ) then ( the person Z has the person Y as uncle ) .}
\langexample{``the plan has failed'' because ``there was a misunderstanding''.}
The first example shows a ``logical rule''; the second example is a ``rationale'' statement.
Parentheses and variables are used to disambiguate. Around 90 grammar rules define the language.
}

\langparax{KANT Controlled English (KCE)}{\pens2551, \prop{t~w~\todo{d?~}a}}{ \cite{mitamura1995tmi} is a controlled natural language for machine translation used within the KANT translation system. The language was first presented under this name in 1995, but it had at that point already been studied and used for several years. The focus is on technical documents, and KCE was the basis for the development of Caterpillar Technical English. Lexicon, grammar, and semantics are restricted. In addition, ambiguities are resolved interactively by augmenting the input sentences with SGML tags. In the following sentence, for example, the attachment of the preposition ``with twelve rivets'' is ambiguous:
\langexample{Secure the gear with twelve rivets.}
In KCE, this ambiguity can be resolved by augmenting the sentence with an SGML tag, for instance ``Secure the gear with $<$attach head=`secure' modi=`with'$>$ twelve rivets.'' For the classification of the language, the question arises whether the SGML tags are part of the language or just a method to keep track of decisions concerning ambiguities. The SGML tags positively contribute to the precision of the language but heavily impede its naturalness. Since such markup tags are usually hidden and since KCE texts are initially written without tags, which are added only afterwards, we consider them a part of the KANT methodology but not of the controlled language itself.}

\langparax{Kodak International Service Language (KISL)}{\pens2551, \prop{c~w~d~i}}{ is a CNL developed at Kodak in the early 1980s.
Some see it as a descendant of CFE \cite{spaggiari2003eamtclaw}.}

\langparax{Lite Natural Language}{\pens5244, \prop{f~w~a}}{ \cite{bernardi2007iwcs} is a CNL based on the language E2V and its variants. It has a deterministic mapping to DL-Lite, which is a logical formalism optimized for good computational properties and is equivalent to a subset of OWL.}

\langparax{``Massachusetts Legislative Drafting Language''}{\pens2551, \prop{c~w~d~g}}{ \cite{massachusetts2003legislative} is a restricted language for legal texts defined by the Massachusetts Senate. Its purpose is ``to promote uniformity in drafting style, and to make the resulting statutes clear, simple and easy to understand and use.'' The language is defined by about 100 rules that restrict syntax (e.g., ``Use the present tense and the indicative mood''), semantics (e.g., ``Do not use `deem' for `consider'{''}), and document structure (``Use short sections or subsections''). In addition, there are close to 90 words and phrases that must not be used, with suggested replacements for each of them (e.g., \emph{hide} instead of \emph{conceal}, and \emph{rest} instead of \emph{remainder}).
}

\langparax{``MILE Query Language''}{\pens4143, \prop{f~w~d~a}}{ \cite{piwek2000inpacts} is a language to access maritime rules and regulations. It follows the conceptual authoring approach in a very similar way as the Drafter and CLEF languages.}

\langparax{Multinational Customized English (MCE)}{\pens2551, \prop{t~w~d~i}}{ \cite{ruffino1982pemt} is a controlled language developed at Xerox to improve the quality of machine-assisted translation. It was based on ILSAM \cite{adriaens1992coling}. It uses a restricted domain-specific vocabulary and ``a set of writing rules which encourage a clear, concise English and a minimization of ambiguities.''}

\langparax{Nortel Standard English (NSE)}{\pens2551, \prop{c~w~d~i}}{ \cite{smart2006claw} is a language developed at Nortel, a telecommunications equipment manufacturer. The development started in 1995 with the help of SMART Communications, and the language was probably influenced by SMART Plain English.
}

\langparax{Naproche CNL}{\pens5333, \prop{f~w~d~a}}{ \cite{cramer2009cnlmain} is a controlled language for mathematical texts similar to CLM and ForTheL. Texts in Naproche CNL can be deterministically mapped to first-order logic and then automatically checked for logical correctness. The following is an excerpt of a proof written in this language:
\langexample{
Axiom 3: For every $x$, $x' \neq 1$.\\
Axiom 4: If $x' = y'$, then $x = y$.\\
Theorem 1: If $x \neq y$ then $x' \neq y'$.\\
Proof: Assume that $x \neq y$ and $x' = y'$. Then by axiom 4, $x = y$. Qed.
}
According to its authors, most texts of mathematical textbooks ``can be rewritten in the Naproche CNL in such a way that they resemble the original text.''
}


\langparax{NCR Fundamental English}{\pens2551, \prop{c~w~d~i}}{ \cite{ncr1978fundeng} is a CNL developed at NCR Corporation.
The language was used for the technical manuals of the company in order to make them ``easier to read and use by NCR employees and customers around the world.'' These are two examples of sentences:
\langexample{While repairing the unit, the field engineer also performs normal maintenance if it is needed.}
\langexample{No maintenance can be performed until the maintenance lock has been activated.}
The language consists of three parts: nomenclature, glossary, and vocabulary. Every word of the language belongs to exactly one of these categories. The nomenclature is an open set of different kinds of named individual entities, such as names of products, tools, routines, as well as named modes and conditions. The glossary is another open set of words for technical concepts, such as \emph{audit trail}, that cannot be replaced by a phrase or brief clause using the words of the vocabulary. The vocabulary, finally, is the most interesting part. It consists of a fixed set of 1,350 words (verbs, nouns, adverbs, adjectives, pronouns, prepositions, articles, and conjunctions) plus 650 abbreviations. The content of the vocabulary ranges from fundamental words such as \emph{a}, \emph{not}, and \emph{in} to domain-specific terms such as \emph{testware}, \emph{calibrate}, and \emph{taxable}. The meaning of these words is restricted, and each comes with a definition in full English. The noun \emph{medium}, for instance, is defined as ``a method of payment'' and must not be used in any other sense. The grammar is not explicitly restricted.
}

\langparax{Oc\'e Controlled English}{\pens2551, \prop{t~w~d~i}}{ \cite{cucchiarini2002oce} is a controlled language developed at Oc\'e, a Dutch company in the printing and copying business.
Oc\'e Controlled English is combined with traditional machine translation techniques to improve the translation quality of the company's documentation in 17 different languages.
One of the important properties of the language is that it leads to more concise texts. For example, instead of ``In several windows, an icon shows the current status/activity of a printer. See the list below for a description of each status.'', one would write:
\langexample{These icons show the status or activity of the copier.}
The language is implemented with the help of the MAXit Checker by SMART Communications. \todo{Influenced by one of the SMART languages?}}

\langparax{OWL ACE}{\pens5243, \prop{f~w~a}}{ \cite{kaljurand2006ppswr} is a controlled language for the ontology language OWL. Syntactically, it is a subset of ACE. Semantically, it is tailored towards the expressiveness of OWL and is more specific than ACE with its underspecified semantics, particularly in the case of plurals. Thus, OWL ACE is more precise but less expressive than ACE.}

\langparax{``OWLPath's Guided English''}{\pens5234, \prop{f~w~a}}{ \cite{garcia2011smc} is a query language for a tool called OWLPath, with which ontologies can be queried. Statements in this language start with the phrase \emph{View any}. These are two examples:
\langexample{View any COMMODITY has\_quoted\_price in BMF.}
\langexample{View any COMPANY whose STOCK\_PRICE.lastTrade is\_greater\_than \$30 and is\_included\_in Dow\_Jones in 2009-04-24.}
These statements are translated into the SPARQL query language. Even though their structure roughly follows English grammar, they cannot be considered valid English sentences.}

\langparax{OWL Simplified English}{\pens5244, \prop{f w a}}{ \cite{power2012cnl} is a controlled language for the Semantic Web. In contrast to most other approaches, there is no real lexicon, neither built-in nor user-defined. Only a very small number of function words are predefined, and users have to list the verbs they intend to use. All other word categories are inferred based on syntactic clues such as capitalization and adjacent words. This is an example (assuming that \emph{governed} and \emph{lives} are listed as verbs):
\langexample{London is capital of a country that is governed by a man that lives in Downing Street.}
}

\langparax{``PathOnt CNL''}{\pens5144, \prop{f~w~a}}{ \cite{kim2005ijmi,namgoong2007iswcaswc} is a controlled language developed for a tool called PathOnt. The tool is multilingual, supporting English and Korean. Statements in this language are deterministically mapped to RDF triples. These are two exemplary sentences:
\langexample{Nam is a student supervised by a professor named Kim.}
\langexample{A received specimen fixed in formalin is a soft tissue mass.}
The language seems to cover only simple existential statements.
}

\langparax{PENG}{\pens5343, \prop{f~w~a}}{ \cite{schwitter2002dexa} is a controlled language whose name stands for ``Processable English.'' It is a rich but unambiguous language that can be automatically translated via discourse representation structures into first-order logic with equality. It is inspired by ACE, and the approach has a strong focus on predictive editing.
These are two examples:
\langexample{Every animal A eats all plants or eats all animals B that are smaller than A and that eat some plants.}
\langexample{While the fox sleeps, the cat chases a bird.}
}

\langparax{PENG-D}{\pens5343, \prop{f~w~a}}{ \cite{schwitter2004altw} is a language derived from PENG, the main difference being that PENG-D builds upon RDF and OWL instead of discourse representation structures.}

\langparax{PENG Light}{\pens5343, \prop{f~w~a}}{ \cite{schwitter2008ai}
is another language derived from PENG. It maps to the TPTP notation for first-order logic.}

\langparax{Perkins Approved Clear English (PACE)}{\pens2551, \prop{c~t~w~d~i}}{ \cite{pym1990tatc} is a controlled language developed at Perkins, a diesel engine manufacturer and now a subsidiary of Caterpillar. The language was introduced in 1980 and was based on ILSAM. The goal was to improve machine-assisted translation. In order to avoid the use of synonyms, PACE comes with a dictionary which has been gradually extended and counted 2,500 entries in 1990, such as ``passage (n): A drilling along which a fluid moves.'' PACE is summarized in ``Ten Rules of Simplified Writing'':
\begin{center}\small
\begin{minipage}[t]{0.46\textwidth}
1. keep sentences short\\
2. omit redundant words\\
3. order the parts of the sentence logically\\
4. do not change constructions in mid\\
\phantom{4.} sentence\\
5. take care with the logic of `and' and `or'
\end{minipage}
\begin{minipage}[t]{0.46\textwidth}
6. avoid elliptical constructions\\
7. do not omit conjunctions or relatives\\
8. adhere to the PACE dictionary\\
9. avoid strings of nouns\\
10. do not use `ing' unless the word\\
\phantom{10.} appears thus in the PACE dictionary
\end{minipage}
\end{center}
The aim of the first five rules is to make the text short and simple, while the last five rules have the somewhat opposing objective to make the text more explicit. This is an example consisting of two PACE sentences:
\langexample{Loosen the pivot fasteners of the dynamo or of the alternator. Loosen also the fasteners of the adjustment link.}
}

\langparax{PERMIS Controlled Natural Language}{\pens5244, \prop{f~w~d~a}}{ \cite{inglesant2008soups}
is a language for expressing access control policies for grid computing environments. It is based on CLOnE with specific extensions for authorization policies:
\langexample{Staff can print on HP Laserjet 1.}
\langexample{I trust David to say who managers are.}
Such statements are mapped to different formal target notations. Each statement follows one of only nine statement patterns.
}

\langparax{``PILLS Language''}{\pens4153, \prop{t~f~d~a}}{ \cite{bouayad2002lrec} is a language for medical information documents used in a system called PILLS. It follows a similar editing approach as the ICONOCLAST language, which was developed a couple of years earlier by the same research group. With the PILLS approach, different types of documents can be automatically generated from a master document and translated into different languages.}

\langparax{Plain Language}{\pens1551, \prop{c~w~g}}{ or \textbf{Plain English} \cite{sec1998plain,plain2011pl} is an initiative by the US government and other organizations. It had its origins in the 1970s with the goal to make official documents easier to understand and less bureaucratic. ``Use pronouns to speak directly to readers'' and ``Avoid double negatives and exceptions to exceptions'' are two exemplary rules. Unlike other such style guides, many of the guideline rules are strict and, with the Plain Writing Act of 2010, US governmental agencies are obliged to comply with them. With the focus being on human understandability and acceptance, documents in Plain Language do not seem to be considerably more precise or simpler from a computational point of view, when compared to full English.}

\langparax{PoliceSpeak}{\pens2132, \prop{c~s~d~g}}{ \cite{johnson2000ecbc} is a language developed to improve police communications of English and French officers at the Channel Tunnel. The goal was to ``make police communications more concise, more predictable, more stable and less ambiguous.'' The project was launched in 1988 and the language was ready in 1992. It has a similar goal and application area as SEASPEAK and the different air traffic control phraseologies.}

\langparax{``PROSPER Controlled English''}{\pens4343, \prop{f~w~d~a}}{ \cite{grover2000claw} is a language for the specification and verification of hardware designs, developed in the late 1990s.
The language is based on a restricted version of a general English grammar. Sentences of the language can be automatically mapped to a certain type of temporal logic. This is an exemplary sentence:
\langexample{If \textsf{sigi} is high and then is low on the next cycle, then \textsf{sigo} is low and after one cycle becomes high and then after one more cycle becomes low.}
Ambiguity is not completely eliminated, but ambiguous sentences can be automatically spotted and reported to the user.}

\langparax{Pseudo Natural Language (PNL)}{\pens5333, \prop{f~w~a}}{ \cite{marchiori2004wi} is a language designed as a user-friendly language for the Semantic Web.
It builds upon RDF and first-order logic, and uses Prolog to calculate inferences. These are two exemplary sentences:
\langexample{JOHN represents the person ``John Smith'' from the company ``http:// www.example.com/staff''.}
\langexample{if IMPLY has as ARGUMENTS X and Y in this order, then X LOGICAL-IMPLY Y.}
Upper-case words such as \emph{JOHN} act as variables that can be instantiated with concrete definitions involving URIs. PNL is unambiguous and has well-defined semantics, but unnatural capitalization mitigates the naturalness of the language. Its structure looks simple at first sight, but rather complex rules have to be applied in order to resolve ambiguous syntax trees.}

\langparax{``Quelo Controlled English''}{\pens4144, \prop{f w a}}{ \cite{franconi2011dl} is a language introduced in 2010 and used in a query interface called Quelo. This is an exemplary query:
\langexample{I am looking for something. It should be equipped with an automatic transmission system and sold by a car dealer. The car dealer should sell a fleet car.}
Following a conceptual authoring approach, users cannot directly edit the sentences, but they can trigger modification actions on the underlying formal representation.}

\langparax{Rabbit}{\pens5244, \prop{f~w~g}}{ \cite{hart2008eswc} is a controlled language for OWL. It has been developed and used by Ordnance Survey, Great Britain's national mapping agency. Rabbit is designed for a specific scenario, in which it is used for the communication between domain experts and ontology engineers to create ontologies. Three types of statements are supported: declarations, axioms, and import statements. These are examples of the first and second type:
\langexample{Sheep is a concept, plural Sheep.}
\langexample{Every River flows into exactly one of River, Lake or Sea.}
The language is quite simple, being defined by a small number of sentence patterns and some modifications thereof.}

\langparax{Restricted English for Constructing Ontologies (RECON)}{\pens5343, \prop{f~w~a~g}}{ \cite{barkmeyer2012nist} is a language to represent facts and rules in an industrial environment, where these facts and rules have a deterministic mapping to first-order logic. This is an exemplary sentence:
\langexample{If any container contains part of a shipment, it contains no other shipment.}
The language is defined by around 200 rules in Backus-Naur form.
}

\langparax{Restricted Natural Language Statements (RNLS)}{\pens3243, \prop{f~w~d~a}}{ \cite{breaux2005pdsn,breaux2008tosem} is a language for policy statements and software engineering goals introduced in 2004. The following are two exemplary RNLS statements:
\langexample{RNLS \#1: The customer will select access codes.}
\langexample{RNLS \#2: The provider will recommend (RNLS \#1) to the customer.}
The second sentence refers to the first one using its identifier \emph{RNLS \#1}. There is a mapping between RNLS and Description Logic, but it is not clear whether this mapping is automated.
}

\langparax{RuleSpeak}{\pens3442, \prop{c~f~w~i}}{ \cite{ross2003businessrule,omg2008sbvr,ross2013businessrules} is a CNL for business rules. The development of the language started in 1985 and it was first presented in 1994. It is very similar to SBVR Structured English, which emerged later. Each RuleSpeak rule belongs to one of eleven ``functional categories'' such as ``computation rule,'' ``inference rule,'' and ``process trigger.'' For each of these categories specific templates are defined. Computation rules, for example, contain the phrase ``must be computed as'' (or simply ``=''). The first of the following two examples is such a computation rule:
\langexample{A product's cost must be computed as the sum of the cost of all its components.}
\langexample{
An order may be accepted only if all of the following are true:\\
- It includes at least one item.\\
- It indicates the customer who is placing it.
}
Sometimes the color codes of SBVR Structured English are adopted to emphasize the different types of the sentence constituents. Like SBVR Structured English, RuleSpeak is linked to the SBVR standard, which provides formal semantics based on second-order logic with Henkin semantics. However, the mapping from RuleSpeak texts to the logical representation is only defined in an informal way. The strict templates considerably simplify the language, but there is no formal grammar that would fully define the language.}

\langparax{SBVR Structured English}{\pens3442, \prop{c~f~w~i}}{
\emph{See Section \ref{sec:cnls}.}
}

\langparax{SEASPEAK}{\pens2132, \prop{c~s~d~g}}{ \cite{strevens1983esp} is an ``International Maritime English'' designed for clear communication among ships and harbors. Its development started in 1981. It is a controlled phraseology similar to PoliceSpeak and the different air traffic control phraseologies.}

\langparax{SMART Controlled English}{\pens2551, \prop{c~t~w~i}}{ \cite{smart2006claw} is a ``more advanced version'' of ASD Simplified Technical English, developed by the company SMART Communications.
It was probably influenced by SMART Plain English, and has been applied to different areas.
This is an excerpt of a document in SMART Controlled English:
\langexample{
When the Quaternary Pump starts operation, the plunger moves inside the chamber. This movement lets the computer calculate and store a position called ``Top Dead Center'' (TDC).
}
The language is implemented in a tool called MAXit Checker, which is able to spot violations of the restrictions of the language.
}

\langparax{SMART Plain English}{\pens2551, \prop{c~w~i}}{, sometimes called \textbf{Plain English Program (PEP)}, is a controlled language developed and used at SMART Communications since the mid 1980s.\footnote{\url{http://www.smartny.com/plainEnglish.htm}} It is based on CFE and was the basis for HELP and the controlled languages at Clark and Rockwell \cite{adriaens1992coling}. As for SMART Controlled English, the tool MAXit Checker can be used to create compliant documents.}

\langparax{``Sowa's syllogisms.''}{\pens5145, \prop{f~w~a}}{
\emph{See Section \ref{sec:cnls}.}
}

\langparax{Special English}{\pens1551, \prop{c~w~s~g}}{ \cite{voa2009wordbook} is a simplified English developed and used by the Voice of America (VOA), the official external broadcast institution of the US government. The language has been used since 1959 and is still used today for news on radio, television, and the web. This makes it the second oldest English-based CNL (after Basic English) and the only one that has been in use for such a long period by the same organization. At the time of its creation, Special English was probably influenced by Basic English. The vocabulary is restricted to about 1,500 words, which have changed over time. Sentences should be short and should be spoken at a slower speed. There are no explicit restrictions on grammar or semantics.
}

\langparax{SQUALL}{\pens5234, \prop{f~w~a}}{ \cite{ferre2012cnl} is a controlled natural language in the area of the Semantic Web to query and update RDF graphs. Sentences in this language are translated into the query language SPARQL, whereby structural ambiguity is resolved based on a few syntactic rules. This is an example:
\langexample{for every publication ?X, ?X has an author ?A and ?A cite-s ?X}
The language is defined by about 50 simple grammar rules.
}

\langparax{Standard Language (SLANG).}{\pens3142, \prop{c~f~w~d~i}}{
\emph{See Section \ref{sec:cnls}.}
}

\langparax{Sun Proof}{\pens2551, \prop{c~t~w~d\todo{?}~i}}{ \cite{wells2002gi} is a controlled language introduced at Sun for their technical documentation. The initial development of the language lasted from 1999 until 2002. The general objective was to write texts that are ``easier to understand and to translate for humans as well as machines'' but with a clear focus on translatability.
Sun Proof is restricted by three sets of guidelines: style guidelines, grammar rules, and terminology. One of the most important rules is the limitation of the sentence length to 25 words. Other rules include semantic restrictions such as using \emph{may} only for granting permission. This is an exemplary sentence:
\langexample{This chapter provides an overview of the standardized solutions that are required to make the transition from IPv4 to IPv6.}
}

\langparax{Sydney OWL Syntax (SOS)}{\pens5243, \prop{f~w~a}}{ \cite{cregan2007owled} is a controlled language introduced in the context of the Semantic Web. It is based on PENG and provides a bidirectional and complete mapping to the ontology language OWL. These are two exemplary sentences:
\langexample{The class adult is fully defined as any person that has at least 20 as an age.}
\langexample{If X has Y as a father then Y is the only father of X.}
}

\langparax{Template Based Natural Language Specification (TBNLS)}{\pens5234, \prop{f~w~d~a~i}}{ \cite{esser2007dx} is a CNL approach for functional tests of control software for passenger vehicles. The language is defined by 15 templates that provide a mapping to propositional logic with temporal relations. This is an exemplary sentence:
\langexample{If \fbox{Button B$_4$ is down ~~~ P$_1$} occurs, then \fbox{Lamp L$_3$ is red ~~~ P$_2$} hold immediately, until \fbox{10 seconds ~~~ T$_1$} elapsed.}
P$_1$ and P$_2$ represent the propositional variables for the respective boxes, and T$_1$ is a time variable.
}

\langparax{ucsCNL}{\pens5244, \prop{f~w~d~a}}{ \cite{barros2011seke} is a controlled natural language for use case specifications in the area of automated software testing. The language is intended to be unambiguous and is defined by a small number of simple grammar rules. There are imperative sentences to describe user actions, as well as declarative statements to describe the system state before and after user actions:
\langexample{After creating a message with 100 characters, go to the drafts folder}
\langexample{The imported media file is a music file}
}

\langparax{Voice Actions}{\pens3142, \prop{f~s~d~i}}{\footnote{\url{http://support.google.com/android/bin/answer.py?hl=en&answer=1715292}} are a CNL for spoken action commands on the Android mobile phone platform. Currently, the language covers twelve informally defined command patterns such as ``map of,'' ``note to self,'' and ``create a calendar event.'' The following is an example:
\langexample{Create a calendar event: Dinner in San Francisco, Saturday at 7:00PM}
These spoken commands can be automatically interpreted and executed by the system.
}

{
\newpage
\twocolumnnew

\acknowledgments

I would like to thank Norbert E. Fuchs, Stefan H\"ofler, Kaarel Kaljurand, Rich Morin, Rolf Schwitter, Simon Spero, and David Whitten for comments on the article and general discussions on the topic. I am also thankful for the responses from Orlando Chiarello, Esra Erdem, Richard Power, Ronald G. Ross, Nestor Rychtyckyj, Donia Scott, Irina Temnikova, and Andries van Renssen to questions about specific languages. In addition, the feedback from Robert Dale, editor-in-chief of the Computational Linguistics journal, anonymous comments from its editorial board, and the anonymous reviews were very helpful to further improve the article. Lastly, I am extremely thankful to James Tierney for working with me on the manuscript to improve grammar and style.


\bibliographystyle{fullname}

\bibliography{cnl}
}

\end{document}